\documentclass[10pt,twocolumn,letterpaper]{article}
\usepackage{cvpr}

\definecolor{cvprblue}{rgb}{0.21,0.49,0.74}
\usepackage[pagebackref,breaklinks,colorlinks,allcolors=cvprblue]{hyperref}

\usepackage{graphicx}
\usepackage{amsmath}
\usepackage{amssymb}
\usepackage{booktabs}

\usepackage{algorithm}
\usepackage{algpseudocode}
\usepackage{amsmath}
\usepackage{amssymb}

\usepackage{multicol}
\usepackage{multirow}
\usepackage{xcolor}

\usepackage[capitalize]{cleveref}
\crefname{section}{Sec.}{Secs.}
\Crefname{section}{Section}{Sections}
\Crefname{table}{Table}{Tables}
\crefname{table}{Tab.}{Tabs.}

\newcommand{\encoder}{\ensuremath{\mathcal{E}}\xspace}  %
\newcommand{\decoder}{\ensuremath{D}\xspace}  %

\newcommand{\armodel}{\ensuremath{p_{\theta}}\xspace} 
\newcommand{\codebook}{\ensuremath{\mathbb{Q}}\xspace}  
\newcommand{\vocabulary}{\ensuremath{\mathbb{V}}\xspace} 
\newcommand{\real}{\ensuremath{\mathbb{R}}\xspace}

\newcommand{\cc}{cluster classifier\xspace}
\newcommand{\CC}{Cluster Classifier\xspace}
\newcommand{\ccshort}{CC\xspace}

\DeclareMathOperator*{\E}{\mathbb{E}}

\usepackage{lineno}
\usepackage{times}  %
\usepackage{helvet}  %
\usepackage{url}  %
\usepackage{graphicx} %
\usepackage{caption} %
\usepackage{algorithm}

\usepackage{algpseudocode}
\usepackage{amsmath}
\usepackage{amssymb}
\usepackage{cleveref}

\usepackage{multirow}
\usepackage{booktabs}
\usepackage{xspace}
\usepackage{colortbl}
\usepackage{subcaption}

\usepackage{newfloat}
\usepackage{listings}
\usepackage{comment}

\usepackage{cuted}
\usepackage{arydshln}
\usepackage{tikz}

\usepackage{calc}

\definecolor{darkgrey}{gray}{0.8}   %
\definecolor{lightgrey}{gray}{0.9} %

\newcommand{\best}[1]{\cellcolor{darkgrey}{#1}}
\newcommand{\sbest}[1]{\cellcolor{lightgrey}{#1}}

\def\paperID{}
\def\confName{CVPR}
\def\confYear{2026}

\title{%
ClusterMark: Towards Robust Watermarking for Autoregressive Image Generators with Visual Token Clustering %
}

\author{Denis Lukovnikov\thanks{Equal contribution}\\
Ruhr University Bochum\\
{\tt\small denis.lukovnikov@rub.de}
\and
Andreas Müller\footnotemark[1]\\
Ruhr University Bochum\\
{\tt\small andreas.mueller-t1x@rub.de}
\and
Erwin Quiring\\
\_fbeta\\
{\tt\small erwin.quiring@fbeta.de}
\and
Asja Fischer\\
Ruhr University Bochum\\
{\tt\small asja.fischer@rub.de}
}

\begin{document}
\maketitle

\begin{abstract}
In-generation watermarking for latent diffusion models has recently shown high robustness in marking generated images for easier detection and attribution. However, its application to autoregressive (AR) image models is underexplored. Autoregressive models generate images by autoregressively predicting a sequence of visual tokens that are then decoded into pixels using a VQ-VAE decoder. Inspired by KGW watermarking for large language models, we examine token-level watermarking schemes that bias the next-token prediction based on prior tokens. We find that a direct transfer of these schemes works in principle, but the detectability of the watermarks decreases considerably under common image perturbations. As a remedy, we propose a watermarking approach based on visual token clustering, which assigns similar tokens to the same set (red or green). We investigate token clustering in a training-free setting, as well as in combination with a more accurate fine-tuned token or cluster predictor. Overall, our experiments show that cluster-based watermarks greatly improve robustness against perturbations and regeneration attacks while preserving image quality, outperforming a set of baselines and concurrent works. Moreover, our methods offer fast verification runtime, comparable to lightweight post-hoc watermarking techniques.
The source code is available at \\
{\fontsize{9.5}{11}\selectfont \url{https://github.com/lukovnikov/ClusterMark}}
\end{abstract}

\vspace{-1em}

\section{Introduction}
\label{sec:intro}

Watermarking of AI-generated content is a key tool to address misuse~\cite{Marchal2024Generative}, to achieve content authenticity, and to maintain the quality of training data by filtering out AI-generated material. 
Besides classic post-hoc watermarking, one line of research explores techniques for embedding watermarks already during generation.
Most research has focused on diffusion models~\cite{Fernandez2023Stable, Wen2023TreeRing, Yang2024GaussianShading}, while autoregressive (AR) image models received less attention.
However, recent progress in this field %
motivates a comprehensive exploration of AR image model watermarking.

AR models for image generation~\cite{yu2022parti,wang2024emu3,yu2024rar,chang2023muse,sun2024LlamaGen,tian2024nspvar,kumbong2025hmar,chang2022maskgit,wang2025tokenbridge,fan2025fluid,tang2025hart} %
generate images by predicting a sequence of tokens in a latent space that is then transformed to pixels using a decoder. %
A common approach is to use discrete tokens which are obtained after a quantization of the latent space. 

In this work, we explore the embedding of watermarks during the autoregressive generation process.
To this end, we consider token-level watermarks which have previously been studied in the context of large language models (LLMs) generating text.
In particular, we adopt the \textit{KGW watermarking} approach proposed by~\citet{kgw}  due to its simplicity and effectiveness. 
This approach builds on the idea of partitioning tokens into red and green sets, then biasing the model to favor green tokens during generation.
This enables watermark detection based on the proportion of green tokens in a given text.

As a baseline, we first examine the direct transfer of KGW-style watermarks. %
Although this approach works in principle, we find that the watermarks lack robustness under common image perturbations. 
This can be explained by the following observations:
To verify the watermark, a given image needs to be mapped back to the tokens that were originally produced by the AR model to generate the image. 
Firstly, the tokens might not be perfectly reconstructed due to a mismatch between the VQ-VAE's image encoding and decoding procedures.
Secondly, image perturbations further amplify the discrepancy between the original and the reconstructed tokens. 
Hence, a naive application of LLM watermarks is not suitable for image generators. 

As a remedy, we introduce a new watermarking approach based on visual token clustering. 
By clustering similar tokens together and building the green and red sets on cluster level instead of token level, the watermark detection robustness improves considerably. 
Even after image perturbations (e.g. blur), the reconstructed tokens are more likely to fall into the same cluster as the originally generated tokens.

We investigate the clustering approach in two variants: (1) a training-free version that only adapts the sampling procedure and (2) in combination with a  trained token or cluster classifier for more robust token reconstruction.

\vspace{0.5em} \noindent
In summary, our contributions are as follows:
\begin{itemize}
    \vspace{0.5em}
    \setlength{\itemsep}{0.5em} %
    \item \textbf{Clustering for KGW-style watermarking for AR image models.} We are among the first to investigate token-level watermarking for AR image models %
    and propose two novel approaches based on token clustering.
    \item \textbf{Robustness analysis.} We 
    conduct an extensive analysis demonstrating the high robustness of our methods against common image perturbations and regeneration attacks. 
    \item \textbf{Extensive evaluation.} We thoroughly evaluate the proposed techniques on different left-to-right autoregressive image models, i.e., LlamaGen~\cite{sun2024LlamaGen} (versions GPT-B and GPT-L) and RAR-XL~\cite{yu2024rar}.
    Our methods achieve state-of-the-art results with low verification runtime, outperforming the most recently developed concurrent in-generation watermarks for AR image generators with available code~\cite{jovanovic2025watermarking,tong2025training}.
\end{itemize}

\section{Preliminaries}
\label{sec:preliminaries}

\subsection{Autoregressive Image Generation}
There are various works~\cite{yu2022parti,wang2024emu3,yu2024rar,chang2023muse,sun2024LlamaGen,tian2024nspvar,kumbong2025hmar,chang2022maskgit,wang2025tokenbridge,fan2025fluid,tang2025hart} that have developed autoregressive methods for image generation with different sampling approaches. 
In the following, we focus on common approaches of decoding discrete tokens from a latent space, that is, on approaches that rely on vector-quantized variational autoencoders (VQ-VAE) for mapping an image to a sequence of tokens.

Here, a color image $x \in \real^{3\times H \times W}$ is first mapped to a latent space using an encoder~$\encoder$, resulting in continuous feature vectors $ \encoder(x)=z \in \real^{d \times h \times w}$. This latent representation is then quantized to visual tokens $q \in \vocabulary^{h \times w}$ based on a codebook.
More precisely, there exists a fixed vocabulary \vocabulary with a corresponding codebook $\codebook \in \real^{|\vocabulary| \times d} $ containing $d$-dimensional vectors. These vectors are used to look up the nearest token to each of the $h \times w$  $d$-dimensional vectors in $z$, obtaining a collection of visual tokens $q$.
The %
tokens %
can be mapped back to an image by first dequantizing them using the same codebook \codebook and then mapping the latent vectors into pixel space using a corresponding decoder $\decoder: \real^{d \times h \times w} \rightarrow \real^{3\times H \times W}$.

An autoregressive image generator models the joint distribution of image tokens $p(q_1, q_2, \dots, q_{h\cdot w} | s)$ given a semantic context vector $s$ (such as a class or text) by decomposing it into $\prod_i^{h\cdot w} \armodel(q_i | q_{1:i-1}, s)$. %
During generation, a sequence of tokens is consecutively generated by sampling from $\armodel$ at every step.
The default setting is to generate an image by producing tokens describing parts of the image, in a left-to-right, top-down scanning order. We consider this setting for our watermarking approach. Note, however, that alternative orderings and parallel decoding~\cite{chang2023muse,tian2024nspvar,chang2022maskgit,kumbong2025hmar} have also been explored. %

\begin{figure*}[ht]
    \centering
    \includegraphics[width=0.8\linewidth]{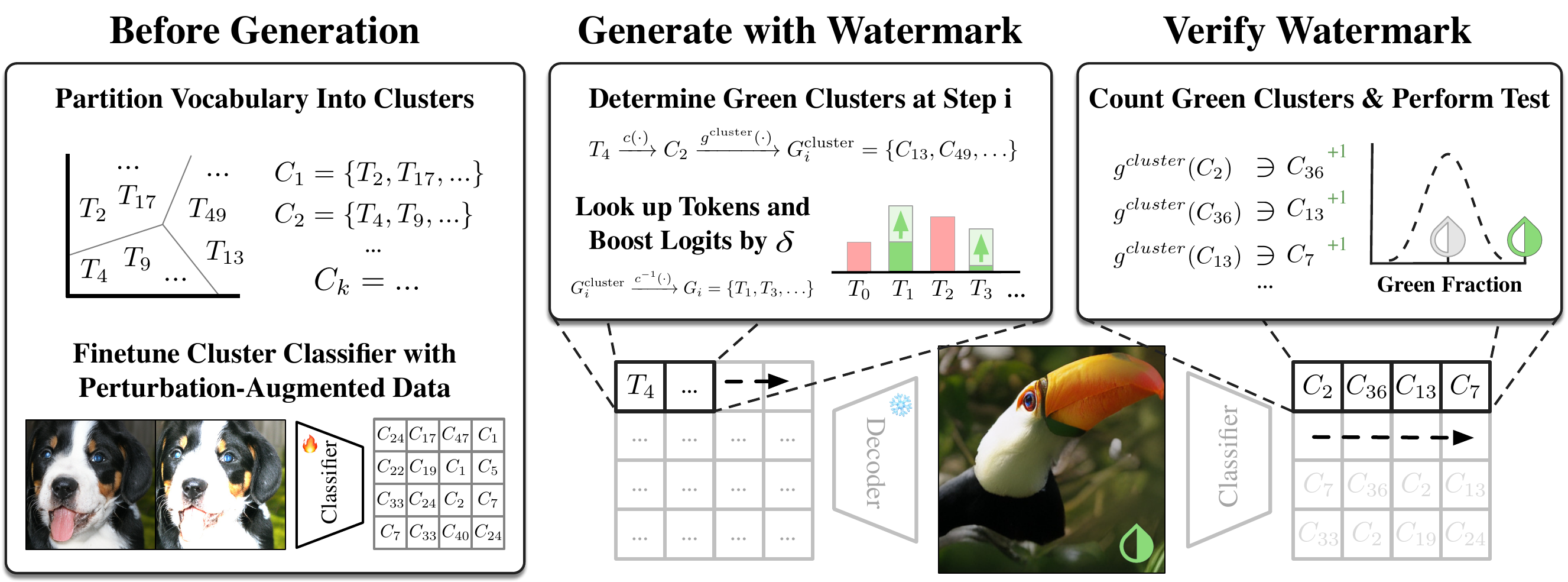}
    \caption{Overview of our proposed cluster-based watermark for AR image generators.
    Before generation, tokens are clustered based on codebook vector similarity, partitioning the vocabulary into a set of clusters. Additionally, a token or cluster classifier can be trained to further boost robustness (This figure shows the variant with the cluster classifer). At every generation step, the set of clusters is partitioned into green and red sets and sampling is biased towards tokens in green clusters. During verification, if the fraction of green tokens is significantly higher than random chance, the image is considered watermarked.
    }
    \label{fig:supp:overview}
\end{figure*}

\subsection{LLM Watermarking}
\label{sec:redgreen}
Several methods have been proposed for watermarking LLM-generated text~\cite{kgw,kirchenbauer2024reliability,kuditipudirobust,christ2024undetectable,dipmark,semamark,yoo2023robust}.
In this work, we adopt the KGW watermarking technique by~\citet{kgw} due to its simplicity and effectiveness.
At every generation step, this technique uses the previously decoded tokens to seed a random function that is used to partition the vocabulary of the generator into two sets, one referred to as ``green", the other as ``red".
More formally, given the sequence of tokens decoded so far, $y_{1:i-1}= ( y_1, \dots, y_{i-1}) \in \vocabulary^{n-1}$, %
we first compute a hash $o_i = \text{hash}(\kappa, y_{i-l}, \dots, y_{i-1})$ using a random hashing function like SHA256.
Here, $\kappa$ is a predefined secret key that remains the same across multiple generations, and $l$ is the number of preceding tokens that are used.
The hash $o_i$ is then used to seed a pseudorandom number generator (PRNG) to randomly select a fraction $\gamma$ of tokens from \vocabulary to form the set of green tokens $G_i$ of size $\lfloor \gamma * |\vocabulary| \rfloor$.
We refer to this process as $g(\cdot)$, which computes the green set based on previously generated tokens: $G_i = g(\kappa, y_{i-l}, \dots, y_{i-1})$.
The red set of tokens is then its complement: $R_i = \vocabulary \setminus G_i$.
Given $G_i$, the bias mask $m_i \in \{0, 1\}^{\vocabulary}$ is defined by setting the mask to one for green tokens: $m_i[v] = \mathbf{1}_{G_i}(v)$.
Next, the model's distribution over \vocabulary is biased to favor tokens from the green set by adding
some value $\delta$ to their logits  $f_{\theta}(y_{1:i-1}) \in \real^{\vocabulary}$ predicted by the model %
before normalizing using the softmax:
$$ p'_{\theta}(y_i | y_{1:i-1}) = \text{softmax} (f_{\theta}(y_{1:i-1}) + m_i * \delta) \enspace .$$
The smaller the value of $\delta$, the less the generator's distribution is affected, but the weaker the watermark signal.

In order to verify the watermark for the given sequence $(y_1, \dots, y_T)$, we can simply recompute the green sets $G_i$ for every given token given the preceding context.
This is fast as there is no need to access the original model.
Finally, to verify the watermark, we perform a one-side (right-tailed) test assessing how unlikely it is to observe the given number of green tokens, $N_g = \#_{\text{green}}$, in the entire sequence by chance.
The null hypothesis is that $N_g$ is distributed according to a binomial distribution: $H_0: N_g \sim \text{Binomial}(T, \gamma)$.
We compute the right-tailed p-value $p = \text{Pr} (X \geq N_g)$ and reject $H_0$ for a sequence if its p-value $p$ is lower than a chosen threshold $\rho$.

\section{Watermarking for AR Image Generators}
\label{sec:approach}
First, we introduce a baseline watermark for autoregressive image generation.
Next, we describe our token clustering and fine-tuning strategies for improving robustness.

\subsection{Baseline AR Watermark}
\label{sec:baseline-variant}

As a \textbf{baseline} variant, we consider the naive application of KGW watermarking 
on the token sequence $\{q_i\}_{1}^{h.w} = (q_1, \dots, q_{h.w})$ produced by the AR image generation model.
The \textbf{generation} of the image consists of two phases: (1) sampling the token sequence $\{q_i\}_{1}^{h.w}$ using the AR model, and (2) mapping the token sequence to an image $x$ using the VQ-VAE.
During sampling, the hash for the red-green partitioning is computed from
the previous token: $o_i = \text{hash} (\kappa, q_{i-1})$.
We refer to $\kappa$ as the (hash) prefix.
For the \textbf{verification} of a given image $x$, we first use the encoder and quantizer of the VQ-VAE to produce a token sequence $\{q'_i\}_{1}^{h.w}$, and then count the fraction of green tokens along the generation order.

\subsection{Cluster-Based AR Watermark}
\label{sec:clustering}
Due to the quantization in the encoding step of the VQ-VAE, even small image perturbations can lead to entirely different tokens being predicted for a large fraction of the tokens (since it is sufficient that the perturbation moves the latent representation closer to another token in the codebook).
This sensitivity of the token reconstruction leads to poor robustness under image perturbations, as similar tokens are randomly assigned to either red or green sets.
Thus, we cluster the codebook tokens based on their embeddings to improve robustness.
Tokens are partitioned into green and red sets such that all tokens belonging to the same cluster are either in the green or in the red set. 

\paragraph{Clustering the vocabulary.}
Prior to generation, we apply $k$-means clustering to partition the vocabulary \vocabulary into $k$ disjoint subsets $(C_1, \dots, C_k)$.
This partitioning is computed using the Euclidean distances between the token representations in the codebook \codebook.
For convenience, we define a cluster assignment function $c(q) \in \{1, \dots, k\}$, which returns the index of the cluster containing token $q$.

\paragraph{Watermarking with clusters.}
During \textbf{generation}, %
there are two main differences compared to the basic case.
The hash $o_i$ is computed based on the cluster of the preceding token: $o_i = \text{hash} (\kappa, c(q_{i-1}))$, where $\kappa$ is the (hash) prefix that is kept constant across multiple generations.
The hash is then used to compute a green set $G_i^{\text{cluster}}$ over clusters $(C_1, \dots, C_k)$, rather than over the vocabulary \vocabulary.
The green set over \vocabulary is simply the union over green clusters: $G_i = \bigcup_{ C_i \in G_i^{\text{cluster}}  } C_i$.
This green set of tokens $G_i$ is then used to bias the generative distribution towards generating one of the green tokens, similarly to the baseline variant.
Note that the red/green assignment is shared by all tokens in a cluster. This should improve robustness to image perturbations, under the assumption that perturbed images are likely to decode to semantically or geometrically similar tokens within the same cluster.
\textbf{Verification} works similarly to the baseline variant with the difference that hashing and masking are again done on cluster level.
\Cref{alg:cluster-watermark} describes the generation and verification process in detail.

\subsection{Token and Cluster Reconstruction}
\label{sec:clusterpredict}
The approach discussed so far %
is training-free and only relies on the clustering of tokens based on their codebook vectors.
As an extension of this cluster-based watermarking approach, we also investigate \textbf{fine-tuning the VAE encoder} to improve cluster reconstructions.
Here, we experiment with two approaches: (1) a token predictor that predicts the token indices that can then be mapped to cluster indices and (2) a cluster predictor that directly outputs the cluster indices. 
To increase robustness, fine-tuning is performed adversarially with perturbation-augmented images by applying a perturbation function $\phi(\cdot)$.

The token and cluster classifier ($\mathcal{M_T}$ and $\mathcal{M}_C$, respectively) are implemented by taking a copy of the VQ-VAE encoder \encoder without the pre-quantization layer, %
and adding a simple classification output layer. %
They are fine-tuned using unwatermarked images generated by the AR model.
The token classifier is trained with the following loss, which tries to produce the original tokens:
\begin{align}
    \mathcal{L}_{\text{TC}} &= \E_{(x,\{q_i\}_1^{h.w}) \sim p_{\theta}} \big[ \sum_i \text{CE}(\mathcal{M}_T\big(\phi(x))_{i}, q_{i}\big) \big] \enspace ,
\end{align}
whereas the cluster classifier is trained with another loss, which tries to reproduce the clusters of the original tokens directly:
\begin{align}
    \mathcal{L}_{\text{\ccshort}} &= \E_{(x,\{q_i\}_1^{h.w}) \sim p_{\theta}} \big[ \sum_i \text{CE}(\mathcal{M}_C\big(\phi(x))_{i}, c(q_{i})\big) \big] ,
\end{align}
Note that the use of these classifiers does not require any changes to the watermarked image generation procedure.%

\begin{algorithm}
\caption{Cluster-Based Watermarking}
\label{alg:cluster-watermark}
\begin{algorithmic}[1]

\Statex
\Function{ComputeGreenSet}{$q_{i-l:i-1}$, \vocabulary, $C$, $\kappa$, $\gamma$}
        \State $\mathcal{G} \leftarrow \text{PRNG}(\text{hash}(\kappa, c(q_{i-l:i-1})))$ \Comment{Seed PRNG}

        \For{$j = 1, 2, \ldots, k$}
            \State $r_j \leftarrow \mathcal{G}.\text{uniform}(0, 1)$
        \EndFor
        \State $k_{\text{green}} \leftarrow \lfloor \gamma \cdot k \rfloor$ \Comment{Number of green clusters}
        \State $G^{\text{cluster}} \leftarrow \text{argsort}((r_1, r_2, \ldots, r_k))[\text{:}k_{\text{green}}]$ 
        \State $G = \bigcup \{ C_j : C_j \in  G^{\text{cluster}} \}$ \Comment{Merge green}
        \State \Return $G$
\EndFunction

\Statex
\Function{GenerateWithWM}{$f_{\theta}$, \vocabulary, $C$, $\kappa$, $\gamma$, $\delta$, $s$}
    \State $q_1 \sim \text{softmax}(f_{\theta}(s)$)  \Comment{Sample first token}
    \For{$i = 2, \ldots, h.w$} \Comment{Generate $h.w$ tokens}
        \State $G_i \leftarrow \text{ComputeGreenSet}(q_{i-l:i-1}, \vocabulary, C, \kappa, \gamma)$
        \State $z_i \leftarrow f_{\theta}(q_{1:i-1}, s)$   \Comment{Compute model logits}
        \For{$v \in G_i$}
            \State $z_i[v] \leftarrow z_i[v] + \delta$   \Comment{Bias green}
        \EndFor
        \State $q_i \sim \text{softmax}(z_i)$ \Comment{Sample next token}
    \EndFor
    \State \Return $q_{1:h.w}$
\EndFunction

\Statex
\Function{VerifyWatermark}{$q_{1:h.w}, \vocabulary, C, \kappa, \gamma$}
    \State $\text{green\_count} \leftarrow 0$
    \For{$i = 2, \ldots, h.w$} %
        \State $G_i \leftarrow \text{ComputeGreenSet}(q_{i-l:i-1}, \vocabulary, C, \kappa, \gamma)$
        \If{$q_i \in G_i$}
            \State $\text{green\_count} \leftarrow \text{green\_count} + 1$
        \EndIf
    \EndFor
    \State $p\text{-value} \leftarrow 1 - \text{CDF}_{\text{Binom}}(\text{green\_count} - 1, h.w, \gamma)$ 
    \State \Return $p\text{-value}$
\EndFunction

\end{algorithmic}
\end{algorithm}

\subsection{Prefix Tuning}
\label{sec:prefixtuning}
In preliminary experiments, we observed that the choice of hash prefix $\kappa$ (the secret) significantly affects watermark detection performance. For certain images (e.g., those with large uniform regions), specific cluster transitions occur more frequently, leading to unusually high green counts even in unwatermarked images and resulting in false positives. To address this, we adopt a simple empirical approach by evaluating multiple $\kappa$ values and selecting the best-performing one.

\section{Evaluation}
\label{sec:experiments}
In this section, we evaluate the effectiveness of our proposed watermarking approach. We first detail the experimental setup, including the models and metrics used, before presenting our main findings and an ablation study.

\subsection{Experimental Setup}
\label{sec:expsetup}

\begin{table*}[t!]
\centering
\resizebox{0.99\linewidth}{!}{%
\begin{tabular}{llcccccccccc}
\toprule
Model &  FID ↓ & Clean & JPEG 20 & Gauss. Blur R 3 & Gauss. std .2 & Salt\&Pepper .1 & Color Jitter & Regeneration \\
\midrule

\multicolumn{10}{c}{LlamaGen GPT-B $256\times256$ (Baseline FID=6.01)}\\
\midrule
DWT-DCT-SVD              & 6.51 & 1.000 | 1.000 & 0.415 | 0.006         & 0.975 | 0.785 & 0.503 | 0.005 & 0.506 | 0.009  & 0.542 | 0.161 & 0.826 | 0.494 \\
RivaGAN                  & 6.34 & 1.000 | 0.999 & 0.928 | 0.510         & \sbest{0.996 | 0.963}  & \sbest{0.951 | 0.661} & \sbest{0.958 | 0.701}  & 0.848 | 0.752 & 0.853 | 0.475 \\
TrustMark & 6.06 & \makebox[\widthof{0.000}][c]{\textcolor{black!50}{N/A}} | 1.000  & \makebox[\widthof{0.000}][c]{\textcolor{black!50}{N/A}} | 0.154 & \makebox[\widthof{0.000}][c]{\textcolor{black!50}{N/A}} | 0.320 & \makebox[\widthof{0.000}][c]{\textcolor{black!50}{N/A}} | 0.023 & \makebox[\widthof{0.000}][c]{\textcolor{black!50}{N/A}} | 0.022 & \makebox[\widthof{0.000}][c]{\textcolor{black!50}{N/A}} | 0.650 & \makebox[\widthof{0.000}][c]{\textcolor{black!50}{N/A}} | 0.373 \\
SSL & 6.19 & 1.000 | 1.000 & 0.837 | 0.242 & 0.994 | 0.938 & 0.553 | 0.035 & 0.545 | 0.022 & 0.982 |0.919 & 0.937 | 0.685 \\
\hdashline
\noalign{\vskip 0.5ex}
IndexMark (+IE)          & 5.84 & 1.000 | 1.000 & 0.969 | 0.821         & 0.761 | 0.171         & 0.631 | 0.055         & 0.635 | 0.071  & 0.907 | 0.732 & 0.951 | 0.761 \\
\hdashline
\noalign{\vskip 0.4ex}
Ours (No Clustering)     & 5.90 & 0.999 | 0.999 & 0.928 | 0.692         & 0.621 | 0.068         & 0.626 | 0.075         & 0.621 | 0.069  & 0.851 | 0.573 & 0.916 | 0.710 \\
+ Token Classifier   & 5.90 & 1.000 | 1.000 & 0.877 | 0.564         & 0.818 | 0.366         & \sbest{0.912 | 0.651} & \best{0.999 | 0.998} & 0.974 | 0.926 & 0.881 | 0.677 \\
Ours (Clustering, k=64)  & 6.12 & 1.000 | 1.000  & \best{0.993 | 0.956} & 0.951 | 0.663         & 0.861 | 0.369         & 0.875 | 0.402        & 0.926 | 0.792 & \best{0.997 | 0.972} \\
+ Token Classifier            & 6.12 & 1.000 | 1.000 & \sbest{0.978 | 0.875}         & \best{0.995 | 0.949}         & \best{0.980 | 0.900} & \best{1.000 | 1.000} & \best{0.960 | 0.917} & \best{0.996 | 0.968} \\
+ Cluster Classifier        & 6.12 & 1.000 | 1.000  & 0.982 | 0.893 & 0.992 | 0.925  & \best{0.982 | 0.895} & \best{1.000 | 0.999}  & \sbest{0.985 | 0.951} & \sbest{0.993 | 0.935} \\

\midrule

\multicolumn{10}{c}{LlamaGen GPT-L $384\times384$ (Baseline FID=4.50)}\\
\midrule
DWT-DCT-SVD              & 4.54 & 1.000 | 0.999 & 0.461 | 0.010        & 0.975 | 0.815         & 0.518 | 0.009        & 0.502 | 0.008         & 0.544 | 0.195 & 0.888 | 0.628 \\
RivaGAN                  & 4.62 & 1.000 | 0.999 & 0.958 | 0.702        & \sbest{0.999 | 0.983}  & 0.950 | 0.682& \sbest{0.959 | 0.733} & 0.859 | 0.790 & 0.910 | 0.596 \\
TrustMark & 4.46 & \makebox[\widthof{0.000}][c]{\textcolor{black!50}{N/A}} | 0.997 & \makebox[\widthof{0.000}][c]{\textcolor{black!50}{N/A}} | 0.338 & \makebox[\widthof{0.000}][c]{\textcolor{black!50}{N/A}} | 0.947 & \makebox[\widthof{0.000}][c]{\textcolor{black!50}{N/A}} | 0.029 & \makebox[\widthof{0.000}][c]{\textcolor{black!50}{N/A}} | 0.025 & \makebox[\widthof{0.000}][c]{\textcolor{black!50}{N/A}} | 0.636 & \makebox[\widthof{0.000}][c]{\textcolor{black!50}{N/A}} | 0.523 \\
SSL & 4.35 & 1.000 | 1.000 & 0.905 | 0.453 & \sbest{0.998 | 0.984} & 0.612 | 0.058 & 0.571 | 0.048 & \sbest{0.988 |0.954} & 0.986 | 0.893 \\
\hdashline
\noalign{\vskip 0.5ex}
IndexMark (+IE)          & 4.49 & 0.999 | 0.999 & 0.976 | 0.836        & 0.841 | 0.307         & 0.642 | 0.035        & 0.654 | 0.041         & 0.929 | 0.793 & 0.983 | 0.834 \\
\hdashline
\noalign{\vskip 0.5ex}
Ours (No Clustering)  & 4.36 & 1.000 | 1.000 & 0.960 | 0.8255        & 0.726 | 0.193         & 0.657 | 0.097        & 0.634 | 0.082         & 0.884 | 0.683 & 0.961 | 0.812 \\
 + Token Classifier     & 4.36 & 1.000 | 1.000 & 0.901 | 0.641        & 0.887 | 0.550         & 0.9128 | 0.698        & \best{0.999 | 1.000} & \sbest{0.980 | 0.954} & 0.929 | 0.729 \\
Ours (Clustering, k=64)  & 4.85 & 1.000 | 1.000   & \best{0.996 | 0.975} & 0.983 | 0.905          & 0.899 | 0.483         & 0.923 | 0.543        & 0.931 | 0.849 & \best{0.999 | 0.997} \\
+ Token Classifier            & 4.85 & 1.000 | 1.000 & 0.985 | 0.931         & \best{0.999 | 0.991}         & \sbest{0.984 | 0.925} & \best{1.000 | 1.000} & \best{0.986 | 0.975} & \sbest{0.999 | 0.991} \\
+ Cluster Classifier      & 4.85 & 1.000 | 1.000  & \sbest{0.989 | 0.944} & \sbest{0.995 | 0.971}  & \best{0.991 | 0.956} & \best{1.000 | 1.000}  & \sbest{0.982 | 0.963}  & \sbest{0.999 | 0.990} \\
\midrule

\multicolumn{10}{c}{RAR-XL $256\times256$ (Baseline FID=3.13)}\\
\midrule
DWT-DCT-SVD             & 3.87 & 1.000 | 1.000 & 0.411 | 0.003        & 0.983 | 0.802 & 0.504 | 0.004        & 0.494 | 0.004         & 0.534 | 0.135 & 0.819 | 0.462 \\
RivaGAN                 & 3.64 & 1.000 | 1.000 & 0.943 | 0.550        & \best{0.998 | 0.983}  & 0.956 | 0.651& \sbest{0.963 | 0.730}  & 0.860 | 0.785 & 0.870 | 0.496 \\
TrustMark & 3.31 & \makebox[\widthof{0.000}][c]{\textcolor{black!50}{N/A}} | 0.998 & \makebox[\widthof{0.000}][c]{\textcolor{black!50}{N/A}} | 0.118 & \makebox[\widthof{0.000}][c]{\textcolor{black!50}{N/A}} | 0.248 & \makebox[\widthof{0.000}][c]{\textcolor{black!50}{N/A}} | 0.024 & \makebox[\widthof{0.000}][c]{\textcolor{black!50}{N/A}} | 0.021 & \makebox[\widthof{0.000}][c]{\textcolor{black!50}{N/A}} | 0.651 & \makebox[\widthof{0.000}][c]{\textcolor{black!50}{N/A}} | 0.350 \\
SSL & 4.31 & 1.000 | 1.000 & 0.855 | 0.227 & \sbest{0.996 | 0.955} & 0.598 | 0.029 & 0.572 | 0.017 & \sbest{0.985 |0.927} & 0.953 | 0.699 \\
\hdashline
\noalign{\vskip 0.5ex}
WMAR (FT+AUGS)          & 4.23 & 1.000 | 1.000 & \sbest{0.992 | 0.954} & 0.964 | 0.652 & 0.879 | 0.476 & 0.861 | 0.423 & 0.914 |0.718 & 0.954 | 0.758 \\
\hdashline
\noalign{\vskip 0.5ex}
Ours (No Clustering) & 3.14 & 1.000 | 0.999 & 0.927 | 0.586 & 0.638 | 0.045 & 0.503 | 0.026 & 0.528 | 0.021 & 0.859 | 0.532 & 0.947 | 0.750 \\
+ Token Classifier & 3.14 & 1.000 | 1.000 & 0.958 | 0.792 & 0.930 | 0.565 & 0.915 | 0.626 & 1.000 | 0.999 & 0.984 | 0.919 & 0.966 | 0.818\\
Ours (Clustering, k=64) & 3.96 & 1.000 | 1.000 & 0.992 | 0.901 & 0.772 | 0.100         & 0.723 | 0.087       & 0.739 | 0.093          & 0.935 | 0.661 & 0.986 | 0.868 \\

+ Token Classifier           & 3.96 & 1.000 | 1.000 & \best{0.995 | 0.960}        & 0.986 | 0.837         & \best{0.984 | 0.885}       & \best{1.000 | 1.000} & \best{ 0.991 | 0.956} & \best{0.994 | 0.945} \\

+ Cluster Classifier       & 3.96 & 1.000 | 1.000 & \best{0.996 | 0.964} & 0.979 | 0.770 &\sbest{0.975 | 0.800} & \best{1.000 | 1.000}   & \best{0.987 | 0.943} & \sbest{0.993 | 0.922} \\
\bottomrule
\end{tabular}

}
\caption{Main results on a set of challenging perturbations. Detailed perturbation configurations are provided in \Cref{sec:supp:perturbation_set_info} of the Supplementary Material. We report AUC and TPR@FPR=1\% across various perturbation and regeneration attacks.
For \emph{Ours (No Clustering)} we set the penalty $\delta=5$ and the green-token fraction $\gamma=0.25$. For \emph{Ours (Clustering)} and for our method using the token/cluster classifier we use $k=64$ clusters and the same $\delta$ and $\gamma$. To improve readability, the \raisebox{0pt}[\dimexpr\ht\strutbox\relax][\dimexpr\dp\strutbox\relax]{\colorbox{darkgrey}{\strut best}} and \raisebox{0pt}[\dimexpr\ht\strutbox\relax][\dimexpr\dp\strutbox\relax]{\colorbox{lightgrey}{\strut second best}} entries in each column are highlighted.}

\label{tab:results:main}
\end{table*}

\paragraph{Models.}
We conduct experiments on three left-to-right autoregressive models: LlamaGen~\cite{sun2024LlamaGen} (GPT-B and GPT-L) and RAR-XL~\cite{yu2024rar}.
For both models, their class-to-image variants are used, which generate images conditioned on ImageNet~\cite{russakovsky2015imagenet} classes.
We use the best reported generation settings.
See Section~\ref{sec:supp:gendetails} in Suppl. Material for details.

\paragraph{Evaluation Metrics.} Similarly to previous work, we report TPR@FPR=1\% and AUC to measure watermark detection performance. The main evaluation (\cref{tab:results:main}) uses 2,000 watermarked images and provides empirical numbers computed against a set of 2,000 generated, unwatermarked images. Image quality is assessed by computing the FID of 50k generated images against the ImageNet validation set.
\vspace{-1em}

\paragraph{Watermark Settings.} We explore different settings for the cluster-based watermark by varying the number of clusters ($k \in {8, 16, 32, 64, 128}$), the strength of the watermark ($\delta \in {2, 5}$), and the fraction of green tokens ($\gamma \in {0.5, 0.25}$).
This is done for different hash prefixes ($\kappa \in {1, \dots, 8}$) from which we pick the best-performing one. The optimal settings were determined on a validation set disjoint from the test set used in \Cref{tab:results:main} and by using different perturbations. This procedure is described in detail in Section~\ref{sec:supp:prefixtune} in the Supplementary Material.

\paragraph{Robustness analysis.}
The robustness is evaluated w.r.t.~both perturbations and regeneration attacks on a distinct set of 2,000 images.
We use an extensive set of image perturbations, such as Gaussian noise and blur.
For regeneration attacks, we use the SD1.5 and FLUX.1 autoencoders, regenerating the watermarked images by first encoding them with the VAE encoder and then decoding them with the VAE decoder. We further add a diffusion-based regeneration attack~\cite{ZhaZhaSu2024invisibleimagewatermarksprovably} with default settings, i.e. 60 steps of denoising using Stable Diffusion 2.1~\cite{RomBlaLor2022stablediffusion}.
Visual examples of all applied perturbations as well as the detailed perturbation settings can be found in Section~\ref{sec:supp:perturbation_examples} in the Supplementary Material.

We additionally evaluated geometric transformations and found that our watermark is vulnerable to rotation and cropping, which is in line with previous in-generation watermarks~\cite{Wen2023TreeRing,Yang2024GaussianShading,Gunn2024Undetectable,jovanovic2025watermarking,tong2025training}. 
This issue can be mitigated by using an \emph{image synchronization layer} such as SyncSeal~\cite{fernandez2025geometric}. A detailed analysis of SyncSeal's ability to undo geometric transformations and restore successful detection of our watermark is provided in Section~\ref{sec:supp:full_geometric_results} in the Supplementary Material.

\paragraph{Baselines.}
We benchmark our AR watermarking approach against post-hoc watermarks: \textit{DWT-DCT-SVD}~\cite{navas2008dwtdctsvd}, \textit{RivaGAN}~\cite{zhang2019rivagan}, \textit{TrustMark}~\cite{bui2025trustmark}, and \textit{SSL}~\cite{fernandez2022sslwatermarking}, applied after generating unwatermarked images.
We also compare against concurrent AR watermarks for which source code is readily available:
(1) \textit{IndexMark}~\citep{tong2025training}, which was implemented only on LlamaGen, and (2) \textit{WMAR}~\citep{jovanovic2025watermarking}, which was developed for RAR-XL among two other models.
For both, we evaluate with their best settings, namely IndexMark with Index Encoder and WMAR with fine-tuning of the VAE encoder and decoder with data augmentation.
Besides these schemes, most other in-generation watermarks target diffusion models and are not applicable to AR models.

\paragraph{Training details.}
To fine-tune the token/\cc, we use 100k images generated using the model \textit{without} applying watermarks.
A predefined set of 64 clusters is used, obtained by k-means clustering.
The model was fine-tuned for 30 epochs with batch size 16, which took $\sim$12 hours on one A40 GPU.
The set of perturbations used during fine-tuning is given in Section~\ref{sec:supp:perturbation_set_info} in the Supp. Material.
We use a linearly increasing perturbation schedule.%

\subsection{Results}

\label{sec:iqvsrobust}
\Cref{tab:results:main} shows our main results. %
The reported watermarking hyper-parameters ($k$ = 64, $\delta$ = 5, $\lambda$ = 0.25) are determined by choosing the best trade-off between image quality (measured using FID) and robustness.
The ablation study in Section~\ref{sec:ablations} gives further details on the selection.
Overall, the results show that cluster-based watermarking indeed significantly improves robustness, even in its training-free variant.
When combined with a token classifier or cluster classifier, it achieves the best overall results in the comparison, also outperforming concurrent work.
Visual examples of generated images are provided in \Cref{fig:minigrid}, with more examples provided in Section~\ref{sec:supp:visual_examples} in the Supplementary Material.

\begin{figure}
    \centering
    \includegraphics[width=1.0\linewidth]{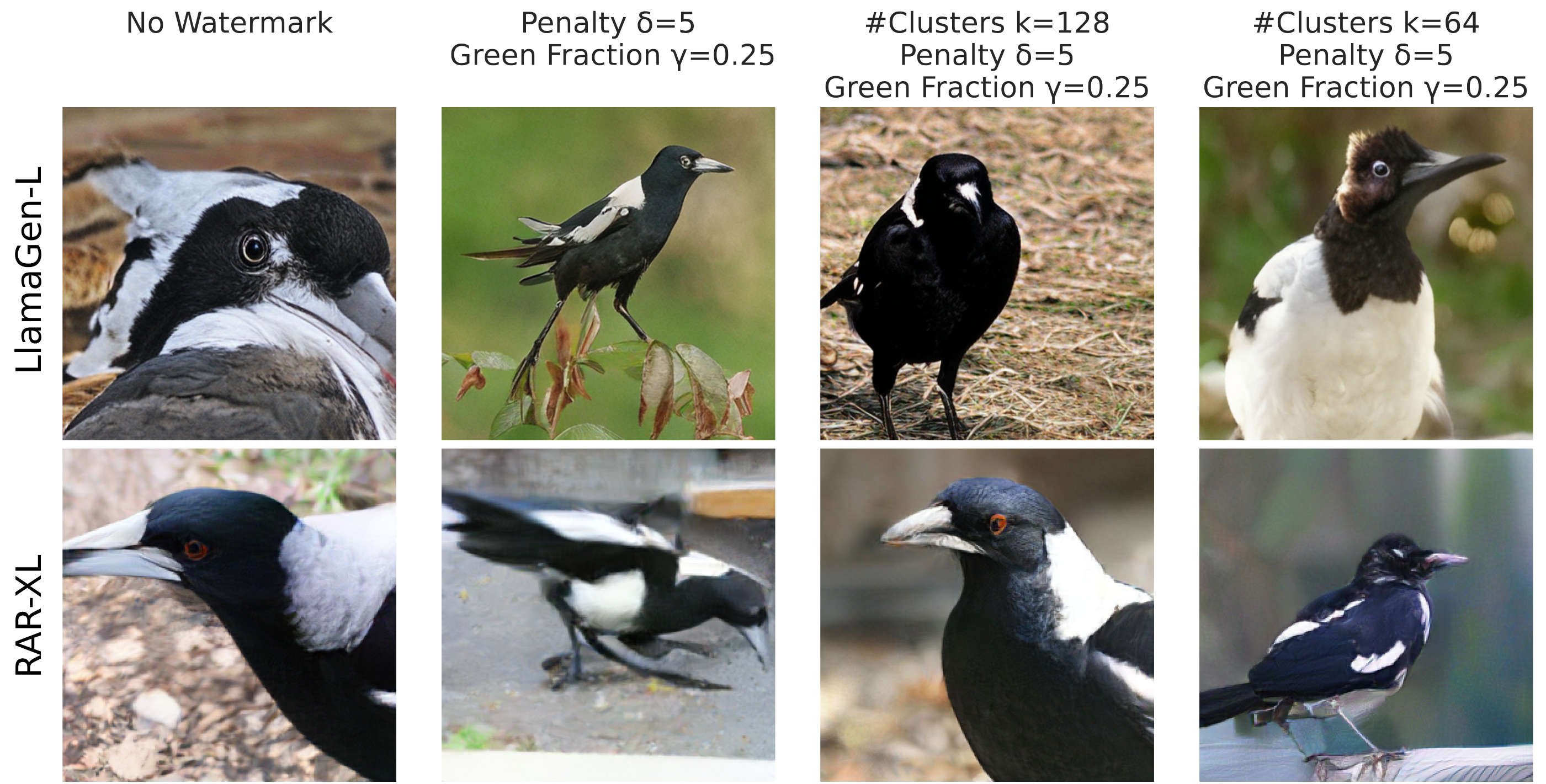}
    \caption{
    Examples of images generated by LlamaGen-L (top) and RAR-XL (bottom), shown (column-wise) (1) unwatermarked images, (2) watermarked images without clustering, and (3,4) watermarked images with clustering enabled.
    Note that while images were generated using the same seed, they differ visually because our watermarking method modifies the generation process rather than applying a post-hoc watermark to given cover images.
    For more examples, see Section~\ref{sec:supp:visual_examples}
    in the Supplementary Material.}
    \label{fig:minigrid}
\end{figure}

\paragraph{Baseline variant.}
Our simple token-level baseline watermarking method (\textit{Ours~(No~clustering)}) shows near perfect TPR@FPR1\% in the clean setting, with image quality matching that of non-watermarked images, as measured by the FID.
However, under even mild image perturbations, such as Gaussian noise, TPR significantly drops, rendering the watermark unusable.

\begin{figure*}
    \centering
    \includegraphics[width=\linewidth]{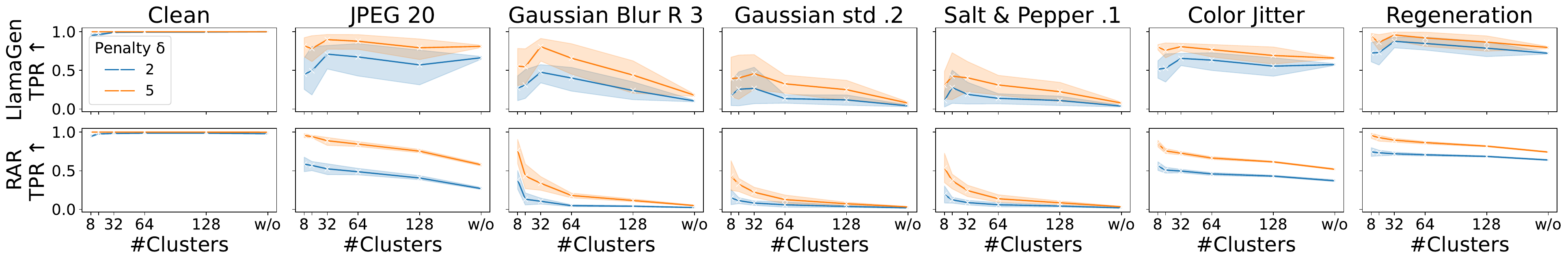}
    \caption{Empirical TPR@FPR=1\% under different perturbations for LlamaGen (GPT-L) and RAR-XL for different numbers of clusters~k and penalties~$\delta$, with green token fraction~$\gamma = 0.25$ for the training-free approach.
    Results are reported over multiple prefixes (8 for RAR-XL and LlamaGen): Lines indicate the average and the shaded area borders the standard deviation. %
    }
    \label{fig:ablation_tpr}
\end{figure*}

\begin{figure}
    \centering
    \includegraphics[width=0.91\linewidth]{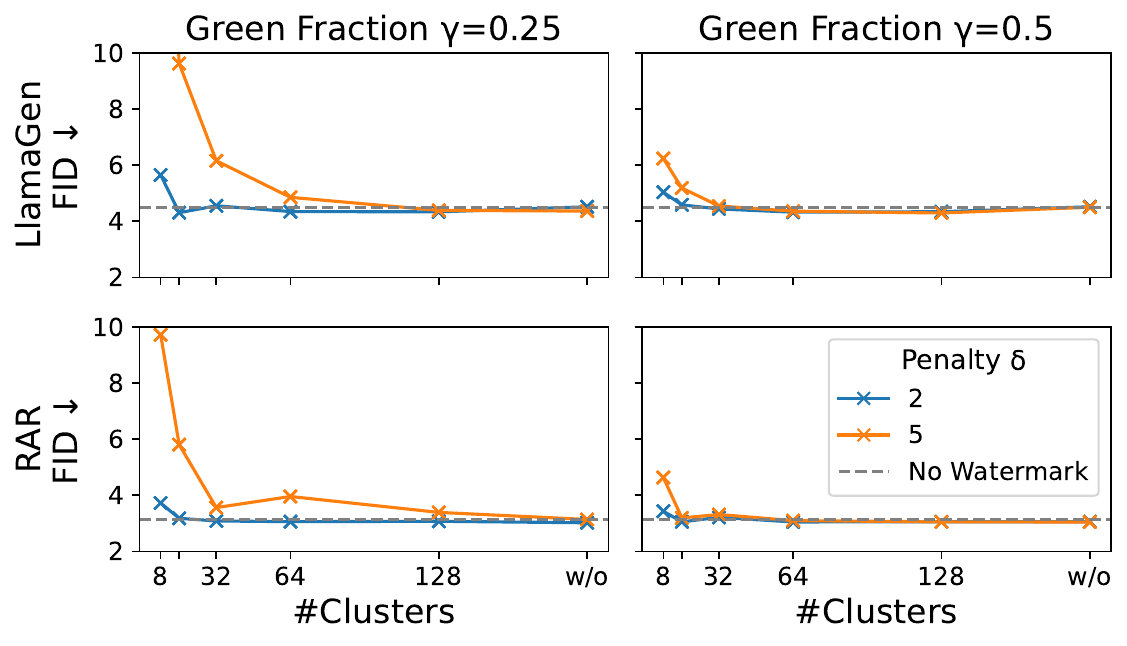}
    \caption{FID obtained for LlamaGen (GPT-L) and RAR-XL across different number of clusters k, different fractions of green tokens~$\gamma$, and penalties~$\delta$.
    See Section~\ref{sec:supp:additional_results} in the Supplementary Material for full results across all settings.
    }
    \label{fig:ablation_fid}
\end{figure}

\paragraph{Effect of clustering.}
The training-free cluster-based approach (\textit{Ours~(Clustering)}) drastically increases robustness against various perturbations for both LlamaGen variants as well as RAR-XL.
While watermark detectability is significantly improved, robustness to blurring and to salt-and-pepper noise remains hard to achieve.
However, note that blurring and salt-and-pepper noise are very noticeable and destructive (see Section~\ref{sec:supp:perturbation_examples} in the Supplementary Material for an example), which severely limits the usefulness of the perturbed image.
With a sufficiently large number of clusters ($k=64$ in \Cref{tab:results:main}), the FID remains low.

\paragraph{Effect of token or cluster classifier.}
Using the \textit{Token Classifier} or \textit{\CC} boosts robustness to several image perturbations for all studied models, outperforming all baselines.
The effect is most pronounced for salt-and-pepper noise, color jitter, and blurring, which are also the most destructive transformations for the training-free variants.
For example, the \cc successfully learned to undo salt-and-pepper noise, yielding near-perfect watermark detection---a clear increase from \mbox{9.3--54.3\%} TPR@1\%FPR when using only clusters ($k=64$).

\paragraph{Comparison to baselines.}
The results show that our best method outperforms all baselines, including the concurrently developed in-generation AR watermarks IndexMark~\cite{tong2025training} and WMAR~\cite{jovanovic2025watermarking}, by a large margin across all tested models on almost all perturbations.
The exception is Gaussian Blur, where RivaGAN achieves slightly higher robustness.
Comparing the FID shows that our methods achieve generator quality close to unwatermarked generation, unlike WMAR, which achieves worse FID in our evaluation using its code and pretrained models.
We also compute CLIP scores and see no significant degradation in text conditioning performance compared to the non-watermarked case. All CLIP scores are in the 29--31 range and are reported in Section~\ref{sec:supp:additional_results} in the Supp. Material.

\begin{table}[t!]
\centering
\resizebox{0.99\linewidth}{!}{%
\begin{tabular}{lrrr}
\toprule
Model & \multicolumn{2}{c}{LlamaGen} & \multicolumn{1}{c}{RAR} \\
Resolution & $256^2$ (GPT-B)  & $384^2$ (GPT-L) & $256^2$ (XL) \\
\midrule

DWT-DCT-SVD & $26.5$~\scriptsize{$\pm11.7$} & $68.4$ \scriptsize{$\pm26.9$} & $23.7$ \scriptsize{$\pm\phantom{0}9.9$} \\
RivaGAN & $24.8$~\scriptsize{$\pm13.6$} & $61.5$~\scriptsize{$\pm20.4$} & $57.6$~\scriptsize{$\pm23.7$} \\
TrustMark & $98.7$~\scriptsize{$\pm15.5$} & $95.2$~\scriptsize{$\pm17.9$} & $95.0$~\scriptsize{$\pm17.9$} \\
SSL & $219.7$~\scriptsize{$\pm\phantom{0}0.4$} & $222.8$~\scriptsize{$\pm\phantom{0}0.9$} & $222.2$~\scriptsize{$\pm\phantom{0}1.1$} \\
\noalign{\vskip 0.3ex}
\hdashline
\noalign{\vskip 0.6ex}

Ours & $12.3$~\scriptsize{$\pm10.9$} & $23.6$~\scriptsize{$\pm10.8$} & $14.9$~\scriptsize{$\pm13.2$} \\
+ Token Classifier & $11.3$~\scriptsize{$\pm\phantom{0}9.3$} & $22.2$~\scriptsize{$\pm10.5$} & $10.5$~\scriptsize{$\pm\phantom{0}8.4$} \\
+ Cluster Classifier & $12.8$~\scriptsize{$\pm10.3$} & $22.7$~\scriptsize{$\pm\phantom{0}8.4$} & $\phantom{0}9.6$~\scriptsize{$\pm\phantom{0}8.6$} \\

\bottomrule
\end{tabular}

}
\caption{
Verification runtime (in milliseconds per image) of our methods against baselines across different models.
}
\label{tab:results:runtime}
\end{table}

\paragraph{Runtime.}
Table~\ref{tab:results:runtime} shows the \textbf{verification runtime} of the methods evaluated in this work.
Time is measured by running watermark verification on single-example batches for 1,000 images.
For the baselines, we use a commonly used implementation for DWT-DCT-SVD and RivaGAN\footnote{\scriptsize\url{https://github.com/ShieldMnt/invisible-watermark}} as well as the official implementations of TrustMark\footnote{\scriptsize\url{https://github.com/adobe/trustmark}} and SSL\footnote{\scriptsize\url{https://github.com/facebookresearch/ssl_watermarking}}, all with GPU acceleration enabled.
The results indicate that the verification of token-based watermarks for AR models is as lightweight as for DWT-DCT-SVD (10-25 milliseconds).
This is in stark contrast to recently proposed in-generation watermarking methods for diffusion models (e.g., Tree-Ring~\cite{Wen2023TreeRing}), which in addition to encoding the image to latent space also require full inversion of the diffusion process, resulting in roughly the same overhead for verification as for generation.
For IndexMark and WMAR, we observe verification times similar to those of our method. For example, WMAR deployed with RAR-XL takes ${\sim}10$ ms (after filling their green lookup table cache), while IndexMark takes $15$ ms (GPT-B) and $30$ ms (GPT-L).

Lastly, the \textbf{generation runtime overhead} is negligible with an average overhead of 0.06s out of 3.7s total generation time and 0.3s out of 14.3s for LlamaGen GPT-B and GPT-L, respectively.

\subsection{Ablation Study}
\label{sec:ablations}
We investigate the effect of the number of clusters $k$ and watermark strength $\delta$ on the TPR in \cref{fig:ablation_tpr}, as well as the effect of $k$, $\delta$ and green token fraction~$\gamma$ on the FID in \cref{fig:ablation_fid}.
Additional ablation details are reported in Section~\ref{sec:supp:additional_results} in the Supplementary Material.

\paragraph{Effect of the number of clusters.}
We tested extremely low numbers of clusters (down to 8) in order to study their effects on image degradation and watermark robustness.
From the TPR plots in~\Cref{fig:ablation_tpr}, we can see that generally, a lower number of clusters results in higher robustness to various perturbations. 
This matches our expectations since fewer and larger clusters result in the reconstructed tokens becoming less likely to leave their original token's cluster after stronger perturbations.
However, we also see a higher variance across different prefixes.
The FID reported in~\Cref{fig:ablation_fid} indicate that with a number of clusters lower than 64, there is a significant drop in generation quality.

\paragraph{Effect of green token fraction~$\gamma$ and penalty~$\delta$.}
Regarding image quality (\cref{fig:ablation_fid}), we observe that $\gamma=0.25$ results in worse FID compared to $\gamma=0.5$, especially when combined with a stronger penalty at extremely low numbers of clusters.
On the other hand, a higher penalty value $\delta$ always results in significantly higher watermark robustness.
A lower green set fraction $\gamma=0.25$ results in significantly higher robustness. This can be seen in \Cref{fig:app:ablation_tpr} in the Supplementary Material.

\paragraph{Effect of hash prefix.}
\label{sec:prefix_effect_analysis}

The prefix $\kappa$ used to seed the pseudo-random number generator for green set generation can have a significant impact on robustness.
As shown in \Cref{fig:ablation_tpr}, lower cluster counts exhibit this effect more strongly, with noticeably higher variance.

To investigate this, we compare the impact of the two hash prefixes which show the best (${\kappa=5}$) and worst (${\kappa=1}$) empirical performance on the green tokens ratio recovered from 2,000 watermarked and unwatermarked images for an extremely low number of clusters ${k=8}$. We use LlamaGen (GPT-B) without the cluster/token-classifier, and set ${\delta=5}$ and ${\gamma=0.25}$. Watermarked images are distorted with the strong perturbation set B to increase the overlap of the two distributions.
This is shown in \Cref{fig:seed_histogram_combined}. 
Ideally, the green token ratio distribution of unwatermarked images should follow a Gaussian centered around ${\gamma=0.25}$. While ${\kappa=5}$ closely approximates this, ${\kappa=1}$ exhibits a heavy right tail that ultimately reduces separability.

This is due to the presence of unwatermarked images with large white areas. 
For some prefixes, green cluster patterns include those transitioning from a flat white area to a nearly identical area (as shown in \cref{fig:seed_histogram_combined} on the right). 
This inflates the green token ratio in unwatermarked images and triggers false positives. 
Lower cluster counts make this behaviour more likely.

\begin{figure}
    \centering
    \includegraphics[width=0.99\linewidth]{figures/prefix_comparison/prefix.pdf}
    \caption{
        Distribution of green token ratios for watermarked and unwatermarked images across 
two hashing prefixes $\kappa$ with a very low number of clusters $k=8$. 
Poorly performing hash prefixes suffer from large uniform regions in unwatermarked 
images, which produce repetitive token bigrams that are spuriously counted as green.
    }
    \label{fig:seed_histogram_combined}
\end{figure}

Prefix tuning is an empirical method to prevent such assignments and can be used to optimize performance during deployment; we leave the exploration of potentially better solutions to future work.

\section{Related Work}
\label{sec:relatedwork}

The integration of watermarks into AI-generated images can be realized in different ways. 
Post-processing techniques~\cite{CoxMilBlo07, alhaj2007dwtdct, Zhu2018hidden, tancik2020stegastamp, zhang2019rivagan, lu2024} apply watermarking to existing images in a post-hoc manner. 
Another possibility is to integrate watermarking directly during the generation process. 
Various techniques have been proposed for text-to-image diffusion models~\cite{Fernandez2023Stable, CiSonYan2024wmadapter, Yang2024GaussianShading, Wen2023TreeRing, CiYanSon24RingID, Gunn2024Undetectable}.
For instance, the decoder of a latent diffusion model (LDM) can be fine-tuned to always create watermarked images from the latents~\cite{Fernandez2023Stable, XioQinFen23Flexible, CiSonYan2024wmadapter}.
Alternatively, inversion-based semantic watermarking modifies the initial latent noise during generation to embed a pattern, which is then recovered through the inversion of the denoising process~\cite{Wen2023TreeRing, Yang2024GaussianShading, CiYanSon24RingID, Gunn2024Undetectable}. 
Since the initial latent is modified, the watermark becomes an inherent part of the generated image. 
In-generation (semantic) watermarking thus exploits the fact that there is a myriad of different ways to generate an image that conforms to user specifications (e.g. the prompt) by partly fixing randomness in the generation procedure.
In this work, we integrate watermarking during generation by biasing the token sampling for AR models for images. %

Concurrently with our work, some recent works have also investigated token-level watermarking for AR image models~\cite{wu2025creweight,jovanovic2025watermarking,tong2025training,kerner2025BitMark}.
\citet{wu2025creweight} propose a different clustering-based approach evaluated on a different model (Emu-3), and only provides a limited robustness study by evaluating robustness to $l_{2}$ and $l_{\infty}$ noise.
\citet{tong2025training} (IndexMark) compute pairs of similar tokens but assign one token in the pair to the green set and the other one to the red set.
We assign similar tokens to the same set.
They also train an index reconstruction model to improve token reconstruction accuracy.
Similarly to \citet{tong2025training}, \citet{jovanovic2025watermarking} (WMAR) also train a token reconstruction model. However, they also consider data augmentation with perturbations during training to improve detection robustness. 
In contrast, our base clustering approach is training-free, simple to use, and improves robustness while preserving image quality.
Our token and cluster classifiers are similar to the trained token reconstructors of IndexMark and WMAR, with a few differences.
WMAR also fine-tunes the VAE decoder, which complicates the training process and further affects the appearance of generated images.
In contrast, our training-based approach is simpler to train and does not require updating the decoder.%

Separate from the detection of AI-generated content, \citet{wang2025safevar} use AR models to embed a hidden image into a given cover image in a post-hoc fashion.
Finally, related to our cluster-based watermarking methods are semantic watermarking methods for LLMs (e.g., \citet{semamark}), developed to resist synonym and paraphrasing attacks by grouping tokens with similar embeddings.

\section{Conclusion}
\label{sec:conclusion}

In this work, we investigate clustering as a promising solution to address the poor robustness of KGW-style watermarking in autoregressive models for image generation.
We thoroughly ablate various design choices across multiple models.
The results indicate that clustering is effective in improving robustness against common perturbations and attacks.
In fact, it significantly boosts watermark detection, except in the most challenging and destructive settings, namely strong salt-and-pepper noise as well as blurring.
In addition to clustering, which is training-free, this work also investigates training token or cluster predictors in combination with clustering. The results indicate that clustering also boosts robustness in this setting, outperforming established post-hoc watermarking baselines and concurrent works on watermarking AR image models.
This work provides a robust alternative for the detection of AI-generated content with low verification runtime overhead, contributing to trustworthy deployment of generative AI.

\newpage

\section*{Acknowledgements}
This work was funded by the Deutsche Forschungsgemeinschaft (DFG, German Research Foundation) under Germany’s Excellence Strategy – EXC 2092 CASA – 390781972 and by the Ministry of Culture and Science of North Rhine-Westphalia as part of the Lamarr Fellow Network.

{
    \small
    \bibliographystyle{ieeenat_fullname}
    \bibliography{bib}
}

\newpage

\appendix
\onecolumn

\section{Full Experimental Setup}
\label{sec:supp:full_experimental_setup}
\vspace{1.5em}

\paragraph{(1) Image Generation Settings.}
\label{sec:supp:gendetails}
\Cref{tab:gendetails} shows detailed generation settings for the models (\textit{LlamaGen}~\cite{sun2024LlamaGen}\footnote{\scriptsize\url{https://github.com/FoundationVision/LlamaGen}} and \textit{RAR-XL}~\cite{yu2024rar}\footnote{\scriptsize\url{https://github.com/bytedance/1d-tokenizer}}) used in this paper. They align with the optimal settings reported in the corresponding original works for each model.

\begin{table}[H]
    \centering
    \begin{tabular}{lccc}
    \toprule
        & \textbf{LlamaGen (GPT-B)} & \textbf{LlamaGen (GPT-L)} & \textbf{RAR-XL} \\
        \midrule
        Resolution & $256\times256$ & $384 \times 384$ & $256\times256$ \\
        $|\vocabulary|$ & 16384 & 16384 & 1024 \\
\hdashline
\noalign{\vskip 0.2ex}
        CFG scale & 2.0 & 2.0 & 6.9 \\
    CFG pow & -- & -- & 1.5 \\
    top-k & $|\vocabulary|$ & $|\vocabulary|$ & $|\vocabulary|$ \\
    top-p & 1.0 & 1.0 & 1.0 \\
    temperature & 1.0 & 1.0 & 1.02 \\
         \bottomrule
    \end{tabular}
    \caption{Model and generation details. $|\vocabulary|$ is the vocabulary size of the VQ-VAE in use.}
    \label{tab:gendetails}
\end{table}

\paragraph{(2) Baseline Settings.}
\label{sec:supp:baseline_details}
The baseline methods \textit{DWT-DCT-SVD}~\cite{navas2008dwtdctsvd} and \textit{RivaGAN}~\cite{zhang2019rivagan} are taken from a commonly used implementation of post-hoc watermarking schemes\footnote{\scriptsize\url{https://github.com/ShieldMnt/invisible-watermark}}. 
For \textit{TrustMark}~\cite{bui2025trustmark} and \textit{SSL Watermarking}~\cite{fernandez2022sslwatermarking} their respective repositories\footnote{\scriptsize\url{https://github.com/adobe/trustmark}}\footnote{\scriptsize\url{https://github.com/facebookresearch/ssl_watermarking}} are used. 
For DWT-DCT-SVD, 64 random bits are encoded for each image. 
For RivaGAN, 32 random bits are encoded for each image as this is the maximum supported by the implementation. Detection is done by extracting bits from images and assessing the ratio of matching bits. For the concurrently developed AR baselines \textit{IndexMark}~\cite{tong2025training} and WMAR~\cite{jovanovic2025watermarking}, we use their respective implementations\footnote{\scriptsize\url{https://github.com/maifoundations/IndexMark}}\footnote{\scriptsize\url{https://github.com/facebookresearch/wmar}}.
For all methods, with the exception of TrustMark, we determine AUC and TPR empirically by calculating ROC against unwatermarked images. For TrustMark, we use their proposed default configuration \textit{Q} which yields an optimal trade-off between quality and watermark robustness. Robustness is evaluated directly by using the built-in detection routine, since decoded bits are only exposed once detection is successful.

\begin{figure*}[t]
\centering
\begin{subfigure}{0.99\linewidth}
    \centering
    \includegraphics[width=\linewidth]{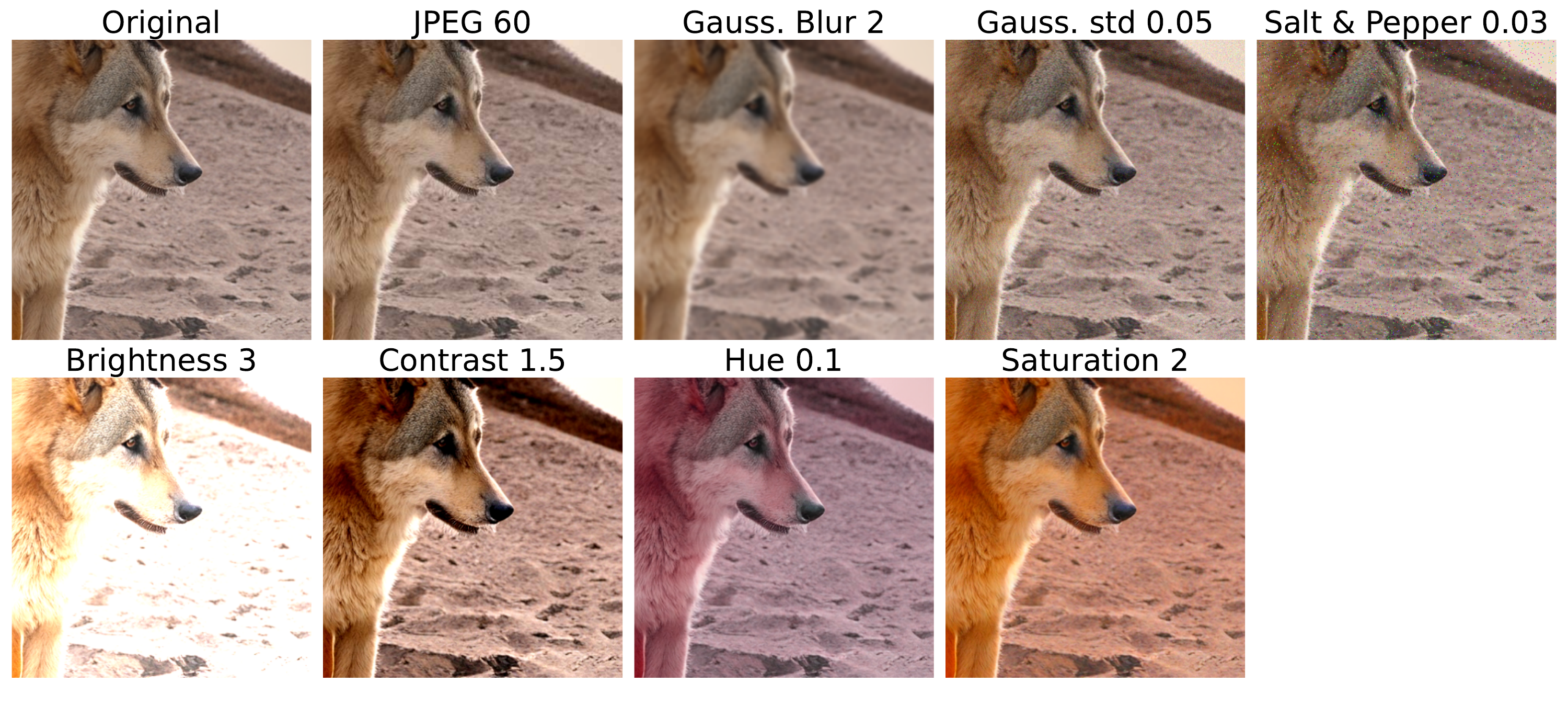}
    \caption{Examples of \textbf{Perturbation Set A} without regeneration attacks (default perturbations used for the ablation and prefix tuning.)}
\end{subfigure}
    
\vspace{0.6em}
    
\begin{subfigure}{0.99\linewidth}
    \centering
    \includegraphics[width=\linewidth]{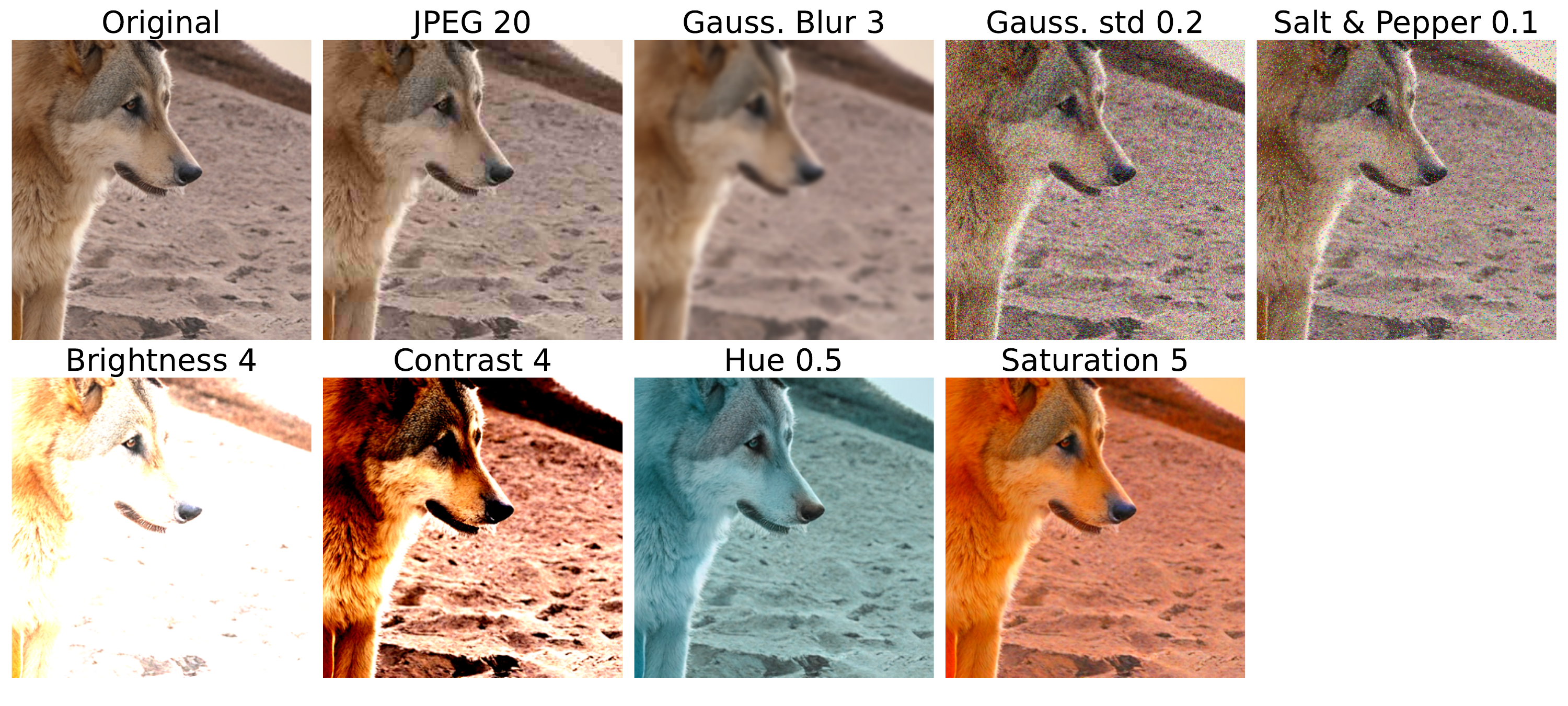}
    \caption{Examples of \textbf{Perturbation Set B} without regeneration attacks (strong perturbations used for main results in Table~\ref{tab:results:main})}
\end{subfigure}
    
\vspace{0.6em}
    
\begin{subfigure}{0.85\linewidth}
    \centering
    \includegraphics[width=\linewidth]{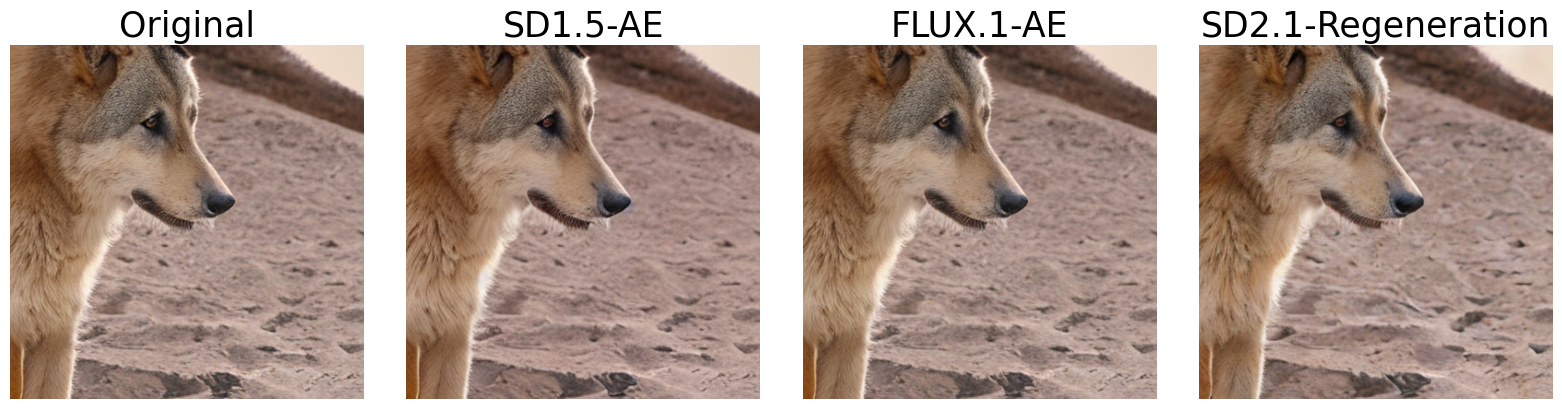}
    \caption{\textbf{Regeneration attacks}. SD1.5-AE and FLUX.1-AE show encode-decode attacks with their respective autoencoders. SD2.1-Regeneration shows the regeneration attack. These attacks are included in both \textbf{Perturbation Sets A and B}.}
\end{subfigure}
\caption{Examples of visual perturbations.}
\label{fig:supp:perturbation_examples}
\end{figure*}

\paragraph{(4) Perturbation Sets and Visual Examples.}
\label{sec:supp:perturbation_set_info}
\label{sec:supp:perturbation_examples}
We use two perturbation sets during evaluation.
\textit{Perturbation Set A} and \textit{Perturbation Set B} are defined in \Cref{tab:supp:merged}.
Set~A includes weaker, more realistic perturbations and has been used in early experiments and for prefix tuning. Set B includes strong perturbations and is used in the main paper to push the watermarking methods to their limits.
Note that for the color jitter perturbations, values are sampled randomly in the range specified by maximum intensity given in the tables.
\Cref{tab:trainperturbset} shows the perturbation settings used during training of the token and cluster classifier.
Figure~\ref{fig:supp:perturbation_examples} shows visual examples of the perturbations applied during all experiments. \textit{Brightness}, \textit{contrast}, \textit{hue}, and \textit{saturation} are summarized as \textit{color jitter}. Results for \textit{SD-1.5-AE}, \textit{FLUX.1-AE} and \textit{SD2.1 Regeneration} are averaged and summarized as \textit{regeneration}. The regeneration attack~\cite{ZhaZhaSu2024invisibleimagewatermarksprovably} is performed with default settings\footnote{\scriptsize\url{https://github.com/XuandongZhao/WatermarkAttacker}}, i.e. 60 steps of denoising using Stable Diffusion 2.1~\cite{RomBlaLor2022stablediffusion}.

\begin{table}[htbp]
\centering
\begin{tabular}{llll}
\toprule
\textbf{Augmentation Type} & \textbf{Parameter} & \textbf{Set A} & \textbf{Set B} \\
\midrule
Gaussian noise            & $\sigma$              & 0.05 & 0.2 \\
JPEG compression          & Quality               & 60   & 20  \\
Gaussian blur             & $r$                   & 2    & 3   \\
RGB salt \& pepper noise  & $p$                   & 0.03 & 0.1 \\
Random Drop               & Ratio                 & 0.3  & 0.5 \\
\hdashline
\noalign{\vskip 0.3ex}
Brightness jitter         & Max. Intensity        & 3    & 4   \\
Contrast jitter           & Max. Intensity        & 1.5  & 4   \\
Hue jitter                & Max. Intensity        & 0.1  & 0.5 \\
Saturation jitter         & Max. Intensity        & 2    & 5   \\
\hdashline
\noalign{\vskip 0.3ex}
SD1.5-AE                  & -                     & -    & -   \\
FLUX.1-AE                 & -                     & -    & -   \\
SD2.1-Regeneration        & Denoising Steps       & 60   & 60  \\
\bottomrule
\end{tabular}
\caption{Perturbation sets A and B. Rows show augmentation types, parameter names and their respective values for both sets.}
\label{tab:supp:merged}
\end{table}

\begin{table}[]
\centering
    \begin{tabular}{ll}
    \toprule
    \textbf{Augmentation Type} & \textbf{Parameter(s)} \\
    \midrule
    Gaussian noise             & $\sigma = 0.1$ \\
    JPEG compression           & Quality range: $80 - 20$ \\
    Gaussian blur              & Max radius $= 2$ \\
    RGB salt \& pepper noise  & Max $p = 0.07$ \\
    \hdashline
    \noalign{\vskip 0.2ex}
    Brightness jitter          & Max. Intensity $= 4$ \\
    Contrast jitter            & Max. Intensity $= 2$ \\
    Hue jitter                 & Max. Intensity $= 0.1$ \\
    Saturation jitter          & Max. Intensity $= 2$ \\
    \bottomrule
    \end{tabular}
\caption{Data augmentation settings and parameters used during training of the token and cluster classifier.}
\label{tab:trainperturbset}
\end{table}

\paragraph{(4) Computing Infrastructrue and Software Details.}
The experiments were conducted on two servers.
\begin{enumerate}
    \item AMD EPYC 7452 (32 cores), $4\times$ Nvidia A40 (48GB), 500 GB RAM, running Ubuntu 20.04.6 LTS
    \item AMD EPYC 7282 (16 cores), $4\times$ Nvidia A40 (48GB), 500 GB RAM, running Ubuntu 20.04.6 LTS
\end{enumerate}
The exact Python version and package requirements can be found in our project repository.

\paragraph{(5) Number of Samples Used in Experiments.}
Images are generated for different purposes: (1) unwatermarked images as negative (unwatermarked) class for evaluation, (2) unwatermarked images used for training the token and cluster classifiers, (3) watermarked images used in robustness study, (4) watermarked images used to measure FID and CLIP score. Below is a breakdown of the number of generated images:

\vspace{5pt}
\textbf{(I) Unwatermarked generated images.} For each of the three models (LlamaGen (GPT-B and GPT-L), RAR-XL), we generated 100k images. 
These images were used for training the cluster and token predictors.
In addition, 2,000 unwatermarked generated images were used as negative class for evaluation.
The baselines methods (DWT-DCT-SVD, RivaGAN, TrustMark, and SSL watermarking) were applied on top of unwatermarked images to obtain positive samples.

\vspace{5pt}
\textbf{(II) Images used for the tuning of watermarking parameters.}
For identifying ideal watermarking parameters, we then generated 2,000 samples for each setting combination (number of clusters $k\in[-, 128, 64, 32, 16, 8]$, penalty $\delta\in[2, 5]$, and green token fraction $\gamma\in[0.5, 0.25]$) for each model. This yields 24 settings per model, totaling $24\times2,000\times3=144,000$ images. This is done for 8 different prefixes $\forall\, \kappa \in \{1, \dots, 8\}$, totaling $144,000\times8=1,152,000$ images. With this data, prefix tuning was then performed.

\vspace{5pt}
\textbf{(III) Images used to evaluate FID and CLIP scores.}
To calculate FID scores and CLIP scores against ImageNet validation set (which contains 50,000 images), 50,000 watermarked images were generated using the best performing hash prefix $\kappa$ obtained in the previous step for each of the 24 settings and for each of the 3 models, yielding $50,000\times24\times3=3,600,000$ images. Since we kept generation seeds consistent over all experiments, the first 2,000 images for each setting are duplicate to the images used in step II.

\textbf{(IV) Images for the main robustness study in Section~\ref{sec:experiments}.}
Since we already obtained 50,0000 watermarked images in the previous step for the best performing hash prefix for each setting, we then sliced 2,000 images which are distinct from the images used in step II for all further evaluation in Section~\ref{sec:experiments}.

\vspace{5pt}
\textbf{Total.} This adds up to $\sim$5M images generated in total. %

\clearpage
\section{Full Experimental Results}
\label{sec:supp:additional_results}

\paragraph{(1) Detailed Results Used of Prefix Tuning.}
\label{sec:supp:prefixtune}
The optimal prefixes $\kappa$ were determined empirically by evaluating 8 possible prefixes $\kappa \in \{1, \dots, 8\}$ across each setup defined by the combination of the generation model, the addition of the token or cluster classifier, the number of clusters ($k \in {8, 16, 32, 64, 128}$), watermark strength ($\delta \in {2, 5}$), and fraction of green tokens ($\gamma \in {0.5, 0.25}$). The evaluation was performed by first applying the weak perturbation set A on 2,000 images generated with the watermark and calculating ROC characteristics against a set of 2,000 unwatermarked images. The best prefix for each group is determined by averaging the results across all perturbations and choosing the maximum TPR@FPR=1\%. These results are shown in Table~\ref{tab:app:full22}. %

To avoid selection bias, the main results shown in Section~\ref{sec:experiments} in the main paper were then performed using the best prefixes, but on a distinct set of 2,000 generated images and using the distinct perturbation set B.

\begin{figure}[H]
    \centering
    \begin{minipage}{0.48\linewidth}
        \centering
        \includegraphics[width=\linewidth]{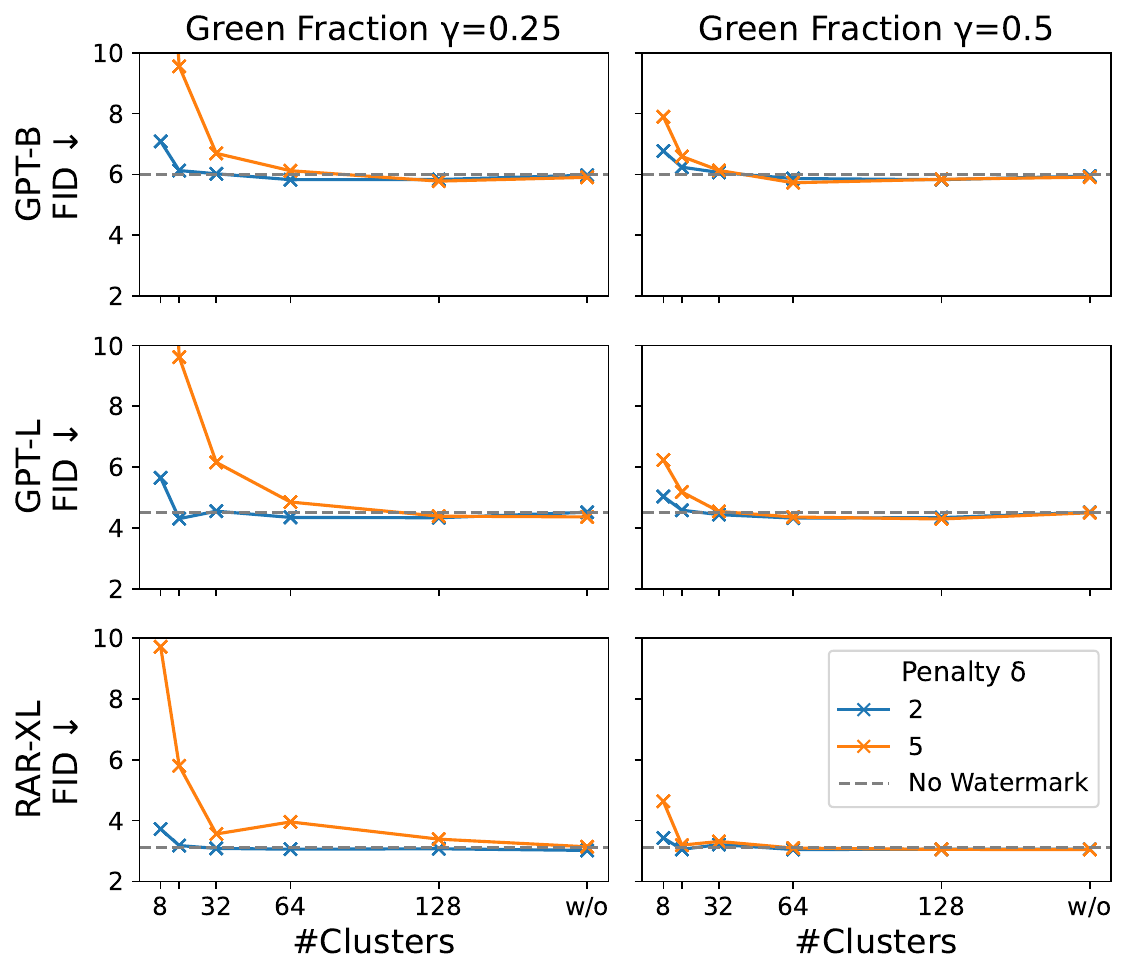}
        \caption{FIDs across models, amount of clusters, penalty $\delta$, and green fraction $\gamma$ settings.}
        \label{fig:app:ablation_fid}
    \end{minipage}%
    \hfill
    \begin{minipage}{0.48\linewidth}
        \centering
        \includegraphics[width=\linewidth]{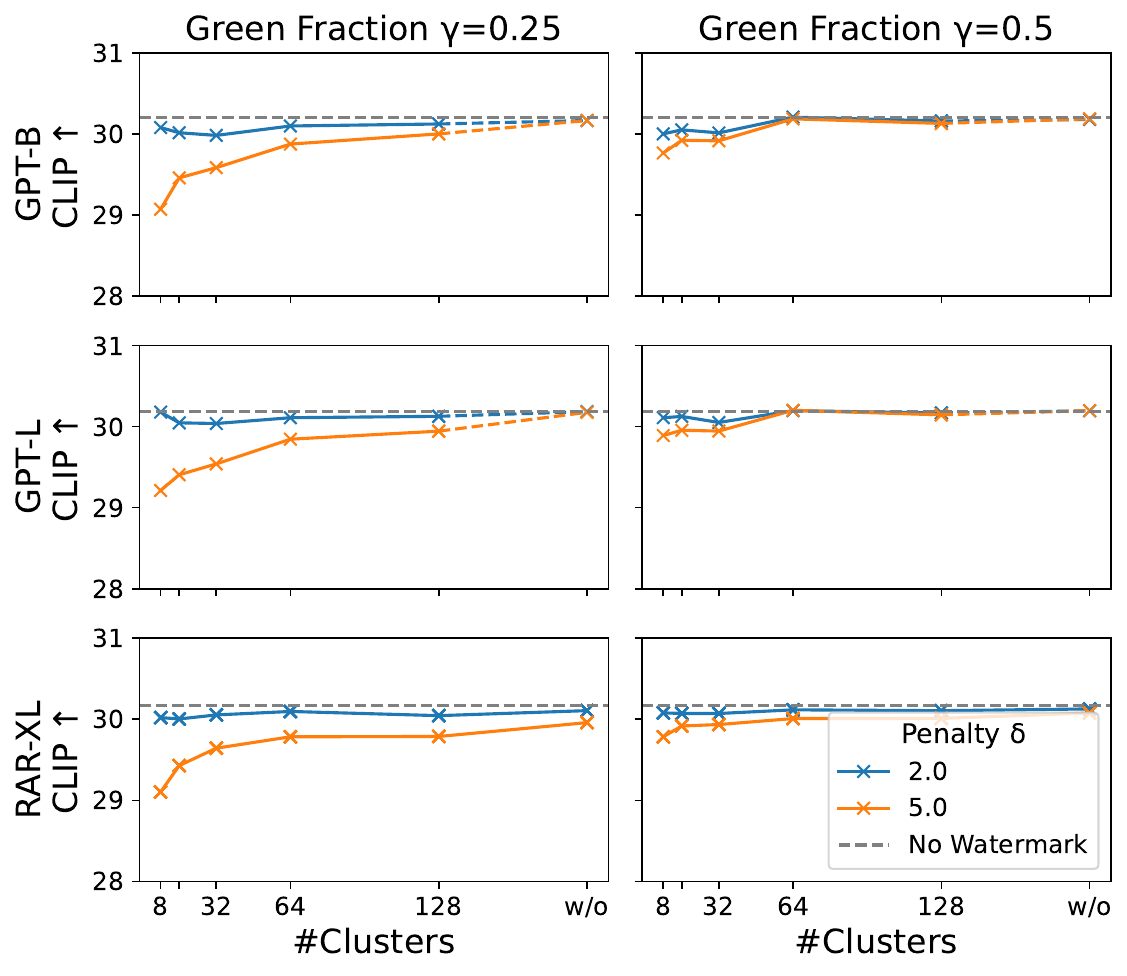}
        \caption{CLIP scores across models, amount of clusters, penalty $\delta$, and green fraction $\gamma$ settings.}
        \label{fig:app:ablation_clip}
    \end{minipage}
\end{figure}

\begin{figure*}
    \centering
    \begin{subfigure}{\linewidth}
        \centering
        \includegraphics[width=1.0\linewidth]{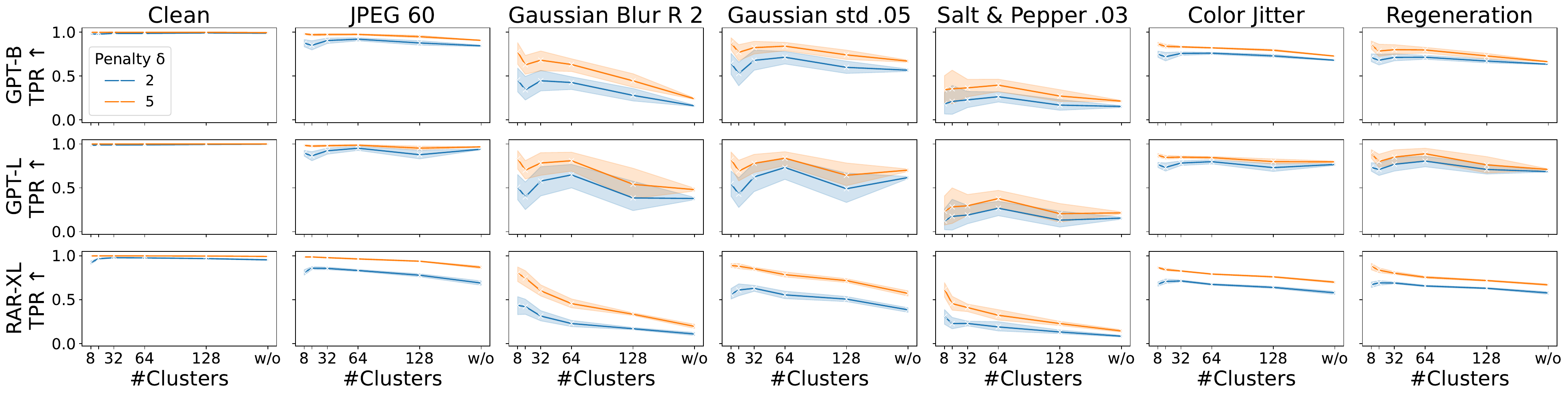}
        \caption{TPR@FPR=1\% for green token fraction $\gamma=0.5$}
        \label{fig:app:ablation_rgfrac_0.5_TPR}
    \end{subfigure}
    
    \vspace{1em} %

    \begin{subfigure}{\linewidth}
        \centering
        \includegraphics[width=1.0\linewidth]{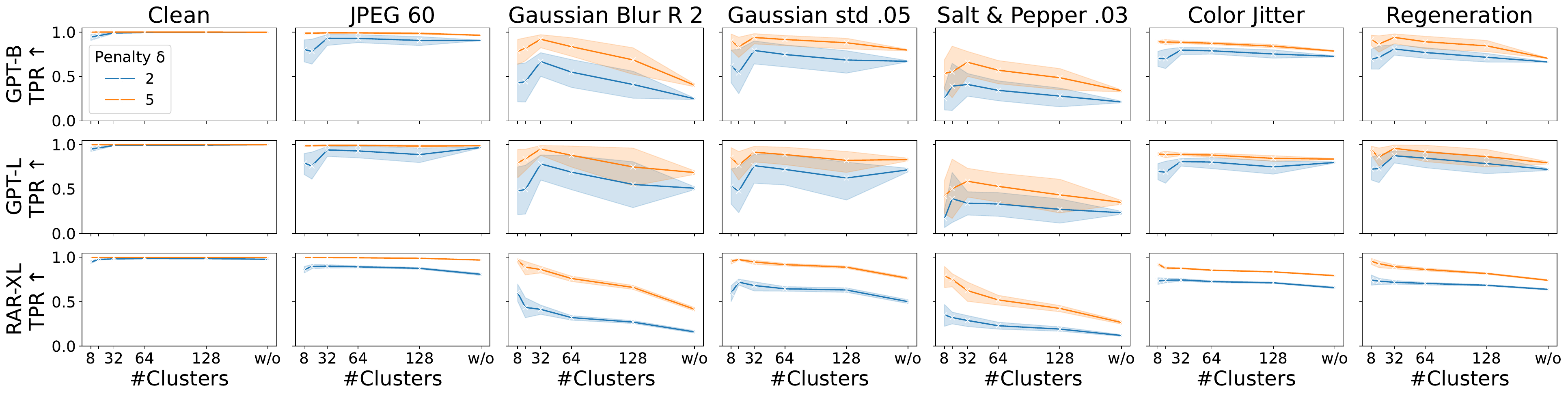}
        \caption{TPR@FPR=1\% for green token fraction $\gamma=0.25$}
        \label{fig:app:ablation_rgfrac_0.25_TPR}
    \end{subfigure}

    \centering
    \begin{subfigure}{\linewidth}
        \centering
        \includegraphics[width=1.0\linewidth]{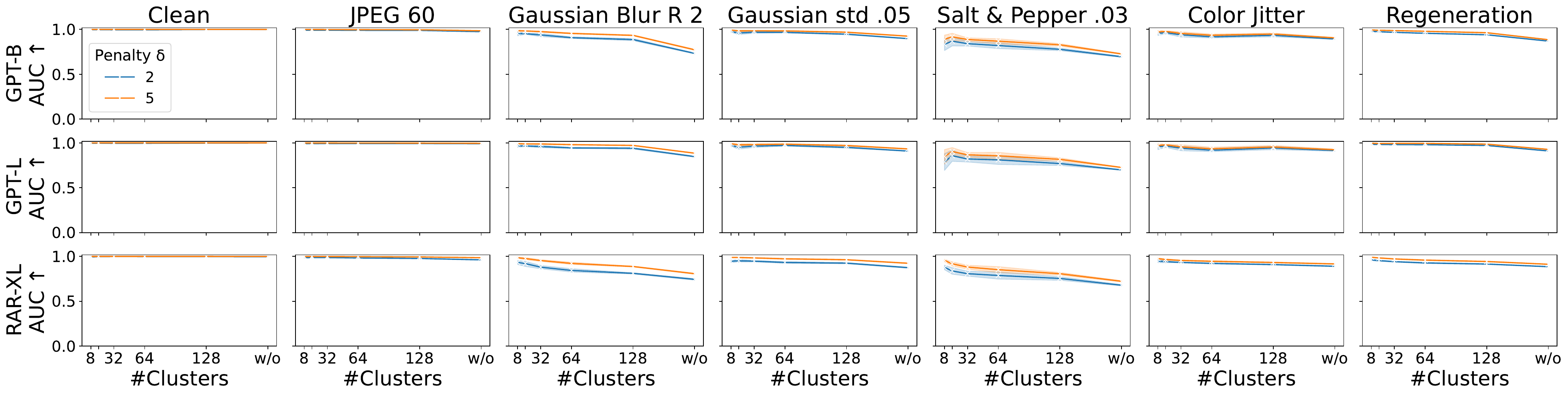}
        \caption{AUC for green token fraction $\gamma=0.5$}
        \label{fig:app:ablation_rgfrac_0.5_AUC}
    \end{subfigure}
    
    \vspace{1em} %

    \begin{subfigure}{\linewidth}
        \centering
        \includegraphics[width=1.0\linewidth]{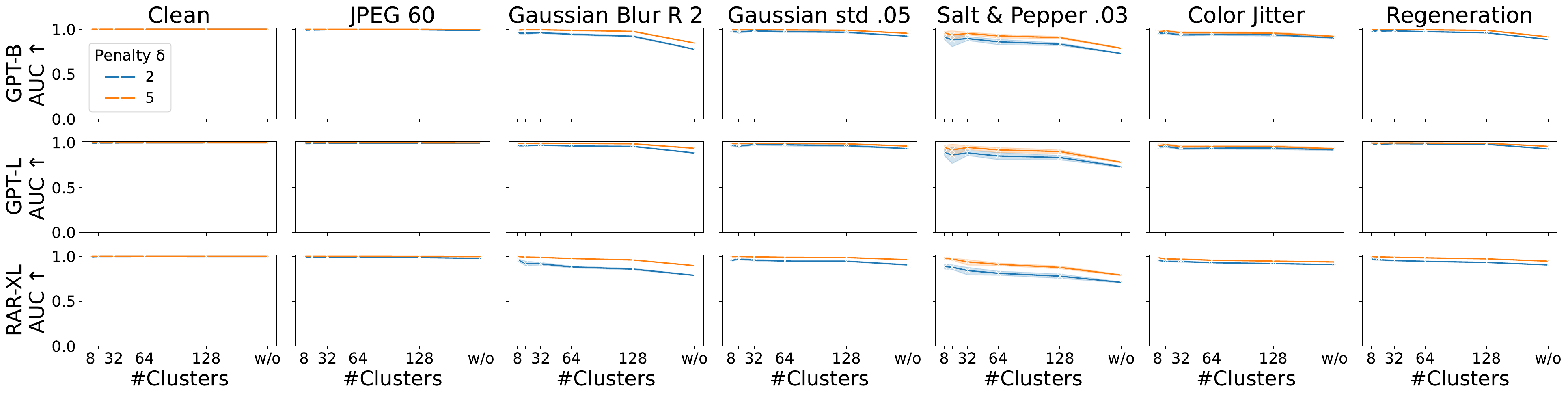}
        \caption{AUC for green token fraction $\gamma=0.25$}
        \label{fig:app:ablation_rgfrac_0.25_AUC}
    \end{subfigure}

    \caption{\textbf{Ablation of detection characteristics} across models, perturbations (Perturbation Set A), amount of clusters, penalty $\delta$, and green token fraction $\gamma$ settings. The results are reported over multiple prefixes (8 for RAR-XL as well as LlamaGen GPT-B and GPT-L), where the standard deviation $\sigma$ is indicated by the colored areas. The lines are averaged over prefixes.
    Note that the \cc is not used here.
    }
    \label{fig:app:ablation_tpr}
\end{figure*}

\paragraph{(2) Full Ablation Plots:} In \Cref{fig:app:ablation_fid}, the full ablation is provided for the effect of different watermarking settings on the FID. 
\Cref{fig:app:ablation_tpr} shows full ablation plots for different perturbations on how watermark detection performs under different watermarking settings. Note however, that the perturbation set used (Set A) is different from Set B used in Figure~\ref{fig:ablation_fid}.

\paragraph{(3) CLIP scores:} We calculate the CLIP scores~\cite{hessel2021clipscore} on 50,000 images across different models and settings by prefixing ImageNet class names corresponding to each generated image class with the prefix "a photo of ". Figure~\ref{fig:app:ablation_clip} shows the results. The CLIP scores are mainly affected by low numbers of clusters.

\paragraph{(4) Results for Geometric Transformations:}
\label{sec:supp:full_geometric_results}
Out-of-the-box, token-level watermarks are not expected to be robust against geometric transformations. In the following, we test their robustness to geometric transformations for our 64-cluster method in \cref{tab:geom,tab:geom:undo,tab:geom:syncseal}.
We then study the effect on the watermark detection by undoing perturbations and by using a synchronization layer. 
Our analysis consists of three parts.

\textbf{First}, we test against various geometric transformations, with results reported in \cref{tab:geom}. As expected, sufficiently strong transformations (x1.5 scaling, 33° rotation, 0.3 perspective, (+6px, +7px) translation, or flipping) essentially destroy the watermark (\textless~10\% TPR, similarly to WMAR without synchronization).
\textbf{Second}, we apply rotation and scaling, and then apply inverse transformations to see how interpolation artifacts affect watermark strength. 
Results are reported in \cref{tab:geom:undo}.
We find that reverted rotation (33°) and scaling (x0.5, x1.5) still lead to 100\% TPR.
\textbf{Finally}, we use SyncSeal~\cite{fernandez2025geometric}, a state-of-the-art image synchronization method, and report results in \cref{tab:geom:syncseal}. %
Just applying SyncSeal on top of our method retains 100\% TPR.
When SyncSeal is used to restore transformed images to their original state after rotation, scaling, flipping or perspective transformations, our watermark (64 clusters, token predictor) still retains 99+\% TPR.
Furthermore, the use of clustering significantly improves TPR compared to the case without clustering.

\paragraph{(5) Effect of Cluster Initialization:}
We also performed a study of the effect of cluster initialization with our 64-cluster variant for RAR-XL.
We computed TPR over 6 different cluster initializations for different perturbations in the stronger perturbation set (Set B). 
We first average over all perturbations in the perturbation set, and subsequently compute the variance over these means.
Without the token predictor, the standard deviation is $0.0226$, while with the token predictor it is $0.0055$.
Table~\ref{tab:clusterinit} reports results for different perturbations in detail.
From the results, we observe that for some perturbations (most notably Gaussian blur), the variance is higher, when performance is lower. This may indicate some sensitivity to cluster initialization, although the lack of prefix tuning possibly also has an impact. The variance is lower when combined with the token predictor.

\begin{table}[ht]
    \centering
    \begin{tabular}{ccccccc}
    \toprule
         &  Clean & G.N 0.2 & JPEG20 & G.Blur R=3 & SP 0.1 & Color Jitter\\
         \midrule
      64 clusters   & $0.999 \pm 0.000$ & $0.115 \pm 0.048$ & $0.866 \pm 0.020$ & $0.296 \pm 0.092$ & $0.129 \pm 0.041$ & $0.692 \pm 0.014$\\
    + token predictor & $1.000 \pm 0.000$ & $0.902 \pm 0.017$ & $0.894 \pm 0.024$ & $0.785 \pm 0.037$ & $1.000 \pm 0.001$ & $0.957 \pm 0.005$ \\
        \bottomrule
    \end{tabular}
    \caption{Effect of cluster initialization on robustness of our method, evaluated on different image perturbations. Reported are the non-empirical TPR@FPR=1\% values with their standard deviation across multiple cluster initializations.}
    \label{tab:clusterinit}
\end{table}

\clearpage

\section{Examples of Generated Images}
\label{sec:supp:visual_examples}
\begin{figure*}
    \centering
    \includegraphics[width=0.99\linewidth]{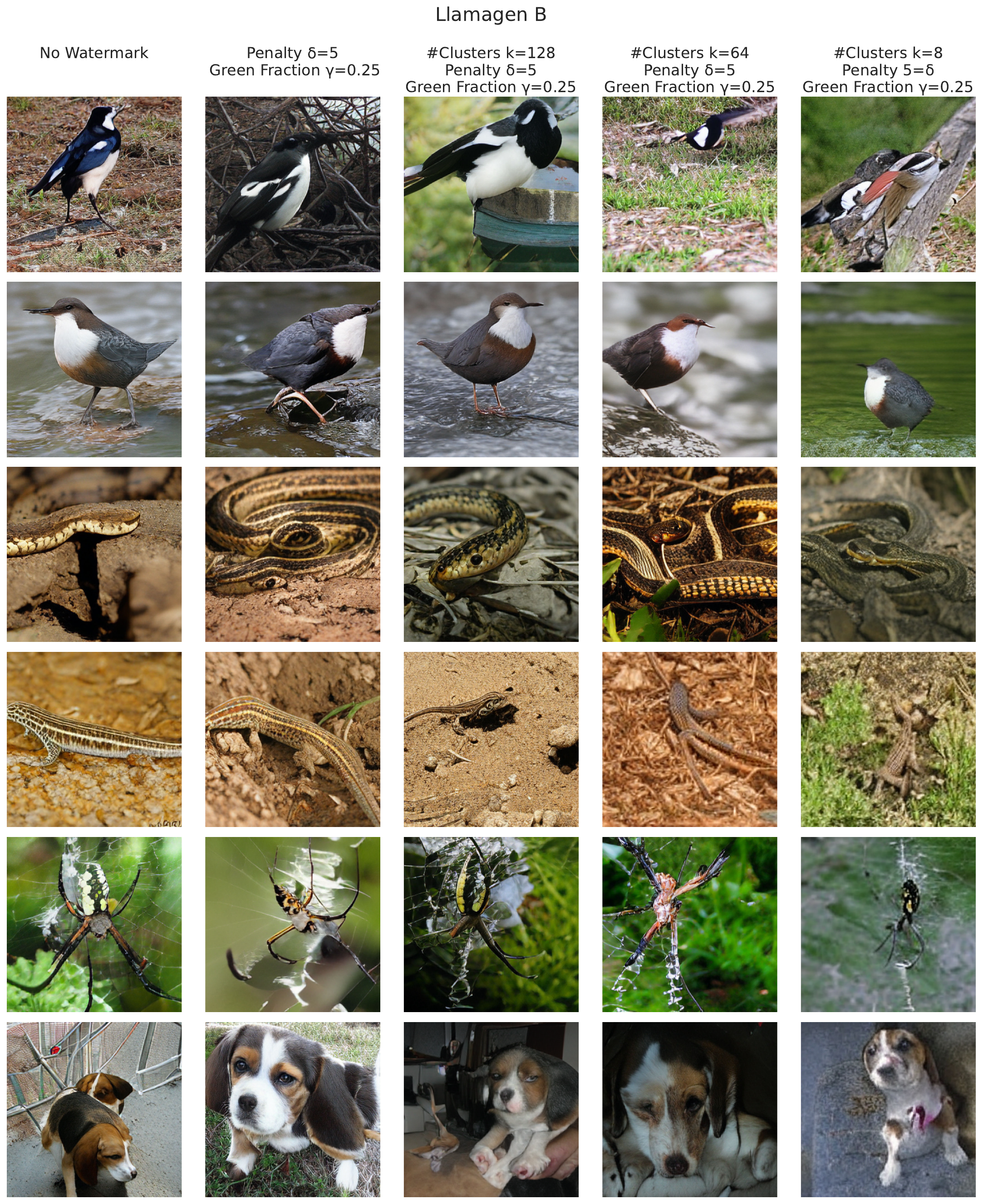}
    \caption{Visual examples of unwatermarked and watermarked images generated with \textbf{LlamaGen (GPT-B)} across different settings.}
    \label{fig:app:example_grid_GPT-B}
\end{figure*}

\begin{figure*}
    \centering
    \includegraphics[width=0.99\linewidth]{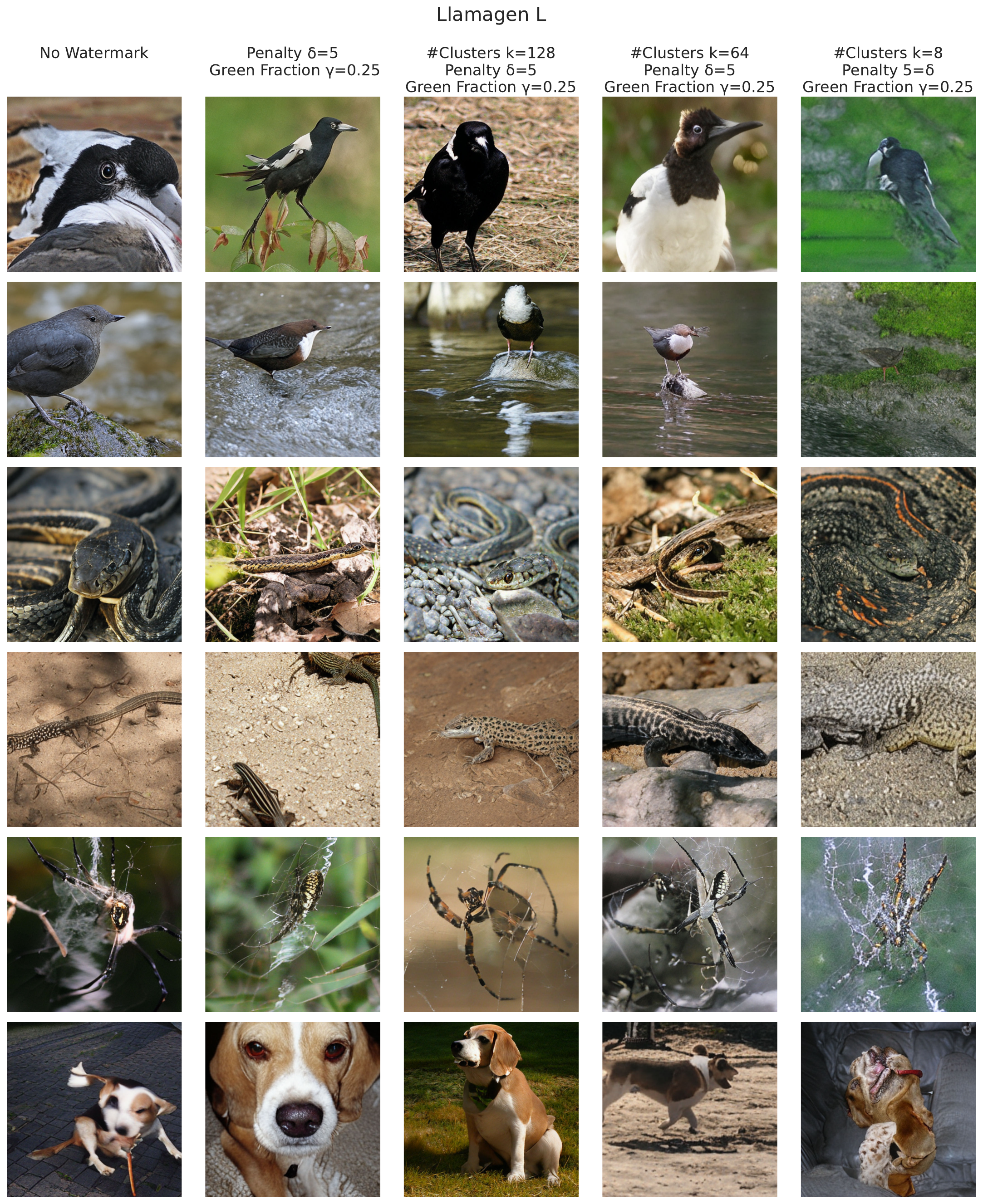}
    \caption{Visual examples of unwatermarked and watermarked images generated with \textbf{LlamaGen (GPT-L)} across different settings.}
    \label{fig:app:example_grid_GPT-L}
\end{figure*}

\begin{figure*}
    \centering
    \includegraphics[width=0.99\linewidth]{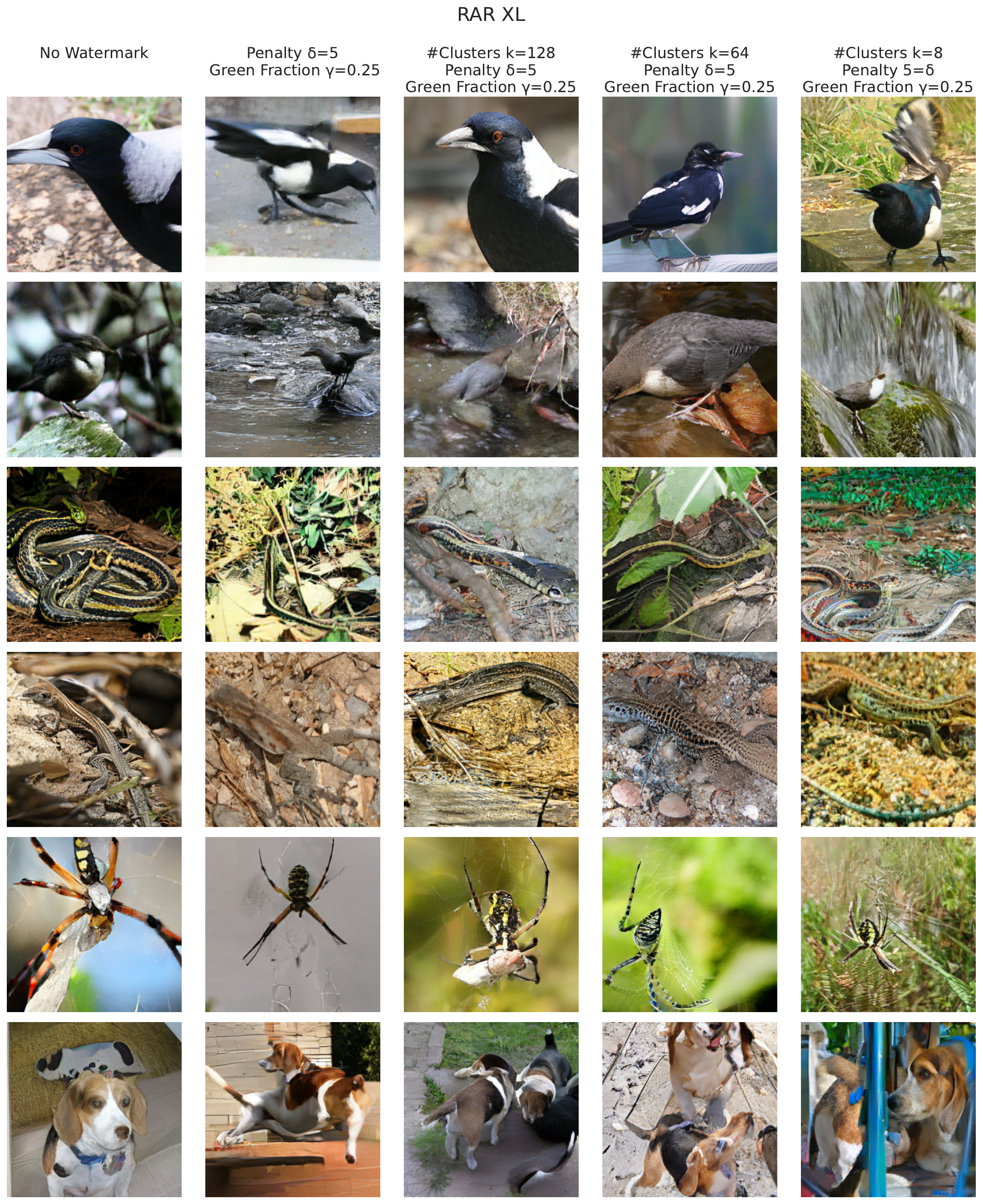}
    \caption{Visual examples of unwatermarked and watermarked images generated with \textbf{RAR-XL} across different settings.}
    \label{fig:app:example_grid_RAR-XL}
\end{figure*}

In \Cref{fig:app:example_grid_GPT-B,fig:app:example_grid_GPT-L,fig:app:example_grid_RAR-XL}, visual examples for both unwatermarked and watermarked images are shown for all models. For watermarked images, we set penalty $\delta=5$ and green token fraction $\gamma=0.25$. We show images generated without clustering ($k=16384$ for LlamaGen GPT-B and GPT-L, $k=1024$ for RAR-XL) as well as images generated with clustering with $k \in [128, 64, 8]$. 
While watermarked images retain image quality down to $k=64$, there is a noticeable decline in image quality for the extreme case of $k=8$.

\section{Security Considerations}
\label{sec:deployment}
In line with previous work on in-generation watermarking~\cite{Yang2024GaussianShading, Wen2023TreeRing, CiYanSon24RingID, Gunn2024Undetectable, jovanovic2025watermarking,tong2025training}, we assume a setting in which a service provider exposes an API for watermarked generation and watermark verification.
The model (AR backbone, VQ-VAE, and token/cluster predictor) must be kept private to help protect against elaborate removal and forgery attacks~\cite{Muller2025semanticforgery, jain2025forgingremoving}.
For example, if the VQ-VAE or token/cluster predictor is leaked, the attacker can learn the token patterns associated with the hashing function.

As we established in Section~\ref{sec:prefix_effect_analysis}, our prefix tuning enables the service provider to rule out unfavourable hash prefixes $\kappa$ empirically. This introduces a potential security concern. As prefix tuning reduces the space of viable prefixes, this could make it easier for an attacker to search for the exact prefix set by the service provider. In our evaluation, we consider a small subset of eight candidates and report the best-performing one. To ensure security in real world deployment, we recommend drawing $\kappa$ from a sufficiently large space (e.g., 64 bits). Even if only a fraction of prefixes are suitable, the resulting effective key space remains large enough that exhaustive search is infeasible. Therefore, restricting $\kappa$ to a subset of viable prefixes does not constitute a security risk in practice. Note that such an attack would additionally require access to the encoder.

\label{sec:supp:fullchallenge}

\begin{table*}[]
\centering
\resizebox{0.92\linewidth}{!}{%
\begin{tabular}{llcccrcccccccccccccc}
\toprule
Model & Method & k & $\delta$ & $\gamma$ & $\kappa$ & FID ↓ & Clean & JPEG 60 & Gauss. Blur R 2 & Gauss. std .05 & Salt\&Pepper .03 & Color Jitter & Regeneration & Average \\
\midrule
\multirow[t]{32}{*}{GPT-B} & No WM & - & - & - & - & $6.01$ & $-$ | $-$ & $-$ | $-$ & $-$ | $-$ & $-$ | $-$ & $-$ | $-$ & $-$ | $-$ & $-$ | $-$ & $-$ | $-$ \\
\cline{2-15}
\noalign{\vskip 0.2ex}
 & \multirow[t]{4}{*}{Ours (No Clustering)} & \multirow[t]{4}{*}{-} & \multirow[t]{2}{*}{2} & 0.50 & 6 & $5.95$ & $0.998$ | $0.989$ & $0.972$ | $0.850$ & $0.739$ | $0.173$ & $0.898$ | $0.582$ & $0.690$ | $0.161$ & $0.896$ | $0.696$ & $0.870$ | $0.640$ & $0.849$ | $0.554$ \\
 &  &  &  & 0.25 & 6 & $5.97$ & $1.000$ | $0.997$ & $0.987$ | $0.909$ & $0.771$ | $0.257$ & $0.920$ | $0.679$ & $0.731$ | $0.237$ & $0.914$ | $0.738$ & $0.879$ | $0.661$ & $0.871$ | $0.608$ \\
\cline{4-15}
\noalign{\vskip 0.2ex}
 &  &  & \multirow[t]{2}{*}{5} & 0.50 & 4 & $5.91$ & $0.999$ | $0.996$ & $0.984$ | $0.913$ & $0.780$ | $0.256$ & $0.926$ | $0.688$ & $0.736$ | $0.229$ & $0.906$ | $0.731$ & $0.886$ | $0.668$ & $0.870$ | $0.606$ \\
 &  &  &  & 0.25 & 4 & $5.90$ & $0.999$ | $0.996$ & $0.995$ | $0.967$ & $0.844$ | $0.426$ & $0.954$ | $0.799$ & $0.783$ | $0.356$ & $0.923$ | $0.794$ & $0.916$ | $0.710$ & $0.898$ | $0.686$ \\
\cline{2-15}
\noalign{\vskip 0.2ex}
 & \multirow[t]{20}{*}{+ Clustering} & \multirow[t]{4}{*}{8} & \multirow[t]{2}{*}{2} & 0.50 & 4 & $6.77$ & $0.990$ | $0.972$ & $0.987$ | $0.942$ & $0.949$ | $0.657$ & $0.972$ | $0.848$ & $0.953$ | $0.558$ & $0.898$ | $0.768$ & $0.974$ | $0.815$ & $0.922$ | $0.739$ \\
 &  &  &  & 0.25 & 5 & $7.09$ & $0.995$ | $0.981$ & $0.992$ | $0.966$ & $0.974$ | $0.794$ & $0.975$ | $0.874$ & $0.932$ | $0.521$ & $0.939$ | $0.825$ & $0.986$ | $0.880$ & $0.948$ | $0.790$ \\
\cline{4-15}
\noalign{\vskip 0.2ex}
 &  &  & \multirow[t]{2}{*}{5} & 0.50 & 4 & $7.90$ & $0.998$ | $0.995$ & $0.999$ | $0.993$ & $0.989$ | $0.914$ & $0.993$ | $0.958$ & $0.982$ | $0.741$ & $0.924$ | $0.839$ & $0.995$ | $0.945$ & $0.949$ | $0.854$ \\
 &  &  &  & 0.25 & 5 & $18.91$ & $1.000$ | $0.999$ & $0.999$ | $0.999$ & $0.998$ | $0.985$ & $0.998$ | $0.983$ & $0.987$ | $0.869$ & $0.977$ | $0.928$ & $0.999$ | $0.995$ & $0.984$ | $0.936$ \\
\cline{3-15}
\noalign{\vskip 0.2ex}
 &  & \multirow[t]{4}{*}{16} & \multirow[t]{2}{*}{2} & 0.50 & 2 & $6.24$ & $0.991$ | $0.978$ & $0.987$ | $0.944$ & $0.944$ | $0.673$ & $0.949$ | $0.741$ & $0.982$ | $0.720$ & $0.931$ | $0.811$ & $0.975$ | $0.824$ & $0.944$ | $0.775$ \\
 &  &  &  & 0.25 & 2 & $6.13$ & $0.997$ | $0.987$ & $0.991$ | $0.965$ & $0.924$ | $0.658$ & $0.979$ | $0.898$ & $0.995$ | $0.942$ & $0.946$ | $0.857$ & $0.983$ | $0.910$ & $0.957$ | $0.852$ \\
\cline{4-15}
\noalign{\vskip 0.2ex}
 &  &  & \multirow[t]{2}{*}{5} & 0.50 & 2 & $6.59$ & $1.000$ | $0.998$ & $0.998$ | $0.988$ & $0.980$ | $0.878$ & $0.980$ | $0.881$ & $0.991$ | $0.864$ & $0.955$ | $0.879$ & $0.996$ | $0.946$ & $0.968$ | $0.882$ \\
 &  &  &  & 0.25 & 8 & $9.56$ & $1.000$ | $1.000$ & $1.000$ | $0.998$ & $0.996$ | $0.967$ & $0.999$ | $0.996$ & $0.996$ | $0.976$ & $0.982$ | $0.947$ & $1.000$ | $0.995$ & $0.988$ | $0.962$ \\
\cline{3-15}
\noalign{\vskip 0.2ex}
 &  & \multirow[t]{4}{*}{32} & \multirow[t]{2}{*}{2} & 0.50 & 6 & $6.07$ & $0.991$ | $0.981$ & $0.990$ | $0.948$ & $0.909$ | $0.576$ & $0.993$ | $0.893$ & $0.888$ | $0.470$ & $0.903$ | $0.787$ & $0.966$ | $0.783$ & $0.917$ | $0.731$ \\
 &  &  &  & 0.25 & 6 & $6.02$ & $0.997$ | $0.994$ & $0.993$ | $0.975$ & $0.949$ | $0.722$ & $0.993$ | $0.947$ & $0.942$ | $0.679$ & $0.942$ | $0.845$ & $0.984$ | $0.889$ & $0.953$ | $0.825$ \\
\cline{4-15}
\noalign{\vskip 0.2ex}
 &  &  & \multirow[t]{2}{*}{5} & 0.50 & 6 & $6.13$ & $1.000$ | $0.999$ & $0.998$ | $0.985$ & $0.967$ | $0.795$ & $0.997$ | $0.956$ & $0.929$ | $0.609$ & $0.932$ | $0.848$ & $0.991$ | $0.890$ & $0.946$ | $0.822$ \\
 &  &  &  & 0.25 & 6 & $6.69$ & $1.000$ | $1.000$ & $0.999$ | $0.996$ & $0.994$ | $0.961$ & $0.999$ | $0.996$ & $0.984$ | $0.903$ & $0.975$ | $0.925$ & $0.999$ | $0.988$ & $0.983$ | $0.939$ \\
\cline{3-15}
\noalign{\vskip 0.2ex}
 &  & \multirow[t]{4}{*}{64} & \multirow[t]{2}{*}{2} & 0.50 & 3 & $5.86$ & $0.993$ | $0.982$ & $0.988$ | $0.940$ & $0.888$ | $0.482$ & $0.979$ | $0.825$ & $0.837$ | $0.364$ & $0.908$ | $0.777$ & $0.958$ | $0.751$ & $0.907$ | $0.690$ \\
 &  &  &  & 0.25 & 7 & $5.83$ & $0.998$ | $0.994$ & $0.995$ | $0.980$ & $0.949$ | $0.740$ & $0.977$ | $0.871$ & $0.901$ | $0.515$ & $0.931$ | $0.827$ & $0.980$ | $0.852$ & $0.939$ | $0.784$ \\
\cline{4-15}
\noalign{\vskip 0.2ex}
 &  &  & \multirow[t]{2}{*}{5} & 0.50 & 3 & $5.73$ & $1.000$ | $0.999$ & $0.998$ | $0.982$ & $0.949$ | $0.685$ & $0.990$ | $0.916$ & $0.890$ | $0.493$ & $0.927$ | $0.828$ & $0.983$ | $0.856$ & $0.935$ | $0.777$ \\
 &  &  &  & 0.25 & 7 & $6.12$ & $1.000$ | $1.000$ & $1.000$ | $0.999$ & $0.992$ | $0.958$ & $0.996$ | $0.974$ & $0.951$ | $0.731$ & $0.958$ | $0.894$ & $0.997$ | $0.972$ & $0.969$ | $0.897$ \\
\cline{3-15}
\noalign{\vskip 0.2ex}
 &  & \multirow[t]{4}{*}{128} & \multirow[t]{2}{*}{2} & 0.50 & 7 & $5.83$ & $0.995$ | $0.987$ & $0.985$ | $0.926$ & $0.897$ | $0.469$ & $0.959$ | $0.718$ & $0.781$ | $0.263$ & $0.905$ | $0.763$ & $0.943$ | $0.711$ & $0.895$ | $0.652$ \\
 &  &  &  & 0.25 & 2 & $5.83$ & $0.999$ | $0.998$ & $0.995$ | $0.969$ & $0.941$ | $0.674$ & $0.969$ | $0.823$ & $0.850$ | $0.407$ & $0.929$ | $0.818$ & $0.965$ | $0.790$ & $0.927$ | $0.742$ \\
\cline{4-15}
\noalign{\vskip 0.2ex}
 &  &  & \multirow[t]{2}{*}{5} & 0.50 & 7 & $5.84$ & $0.999$ | $0.998$ & $0.996$ | $0.970$ & $0.944$ | $0.636$ & $0.977$ | $0.838$ & $0.846$ | $0.381$ & $0.922$ | $0.811$ & $0.967$ | $0.779$ & $0.922$ | $0.731$ \\
 &  &  &  & 0.25 & 2 & $5.78$ & $1.000$ | $1.000$ & $1.000$ | $0.995$ & $0.988$ | $0.915$ & $0.992$ | $0.943$ & $0.911$ | $0.597$ & $0.950$ | $0.871$ & $0.990$ | $0.917$ & $0.957$ | $0.851$ \\
\cline{2-15}
\noalign{\vskip 0.2ex}
 & \multirow[t]{4}{*}{+ Cluster Classifier} & \multirow[t]{4}{*}{64} & \multirow[t]{2}{*}{2} & 0.50 & 2 & $5.86$ & $0.992$ | $0.980$ & $0.974$ | $0.857$ & $0.977$ | $0.924$ & $0.974$ | $0.937$ & $0.990$ | $0.978$ & $0.967$ | $0.930$ & $0.933$ | $0.690$ & $0.964$ | $0.889$ \\
 &  &  &  & 0.25 & 7 & $5.83$ & $0.998$ | $0.992$ & $0.988$ | $0.921$ & $0.992$ | $0.965$ & $0.989$ | $0.961$ & $0.996$ | $0.989$ & $0.985$ | $0.959$ & $0.964$ | $0.779$ & $0.983$ | $0.931$ \\
\cline{4-15}
\noalign{\vskip 0.2ex}
 &  &  & \multirow[t]{2}{*}{5} & 0.50 & 2 & $5.73$ & $1.000$ | $1.000$ & $0.993$ | $0.953$ & $0.997$ | $0.984$ & $0.994$ | $0.982$ & $1.000$ | $1.000$ & $0.983$ | $0.965$ & $0.966$ | $0.777$ & $0.984$ | $0.943$ \\
 &  &  &  & 0.25 & 7 & $6.12$ & $1.000$ | $1.000$ & $0.998$ | $0.992$ & $0.999$ | $0.998$ & $0.999$ | $0.996$ & $1.000$ | $1.000$ & $0.993$ | $0.985$ & $0.993$ | $0.935$ & $0.995$ | $0.981$ \\
\cline{1-15}
\noalign{\vskip 0.2ex}
\multirow[t]{32}{*}{GPT-L} & No WM & - & - & - & - & $4.50$ & $-$ | $-$ & $-$ | $-$ & $-$ | $-$ & $-$ | $-$ & $-$ | $-$ & $-$ | $-$ & $-$ | $-$ & $-$ | $-$ \\
\cline{2-15}
\noalign{\vskip 0.2ex}
 & \multirow[t]{4}{*}{Ours (No Clustering)} & \multirow[t]{4}{*}{-} & \multirow[t]{2}{*}{2} & 0.50 & 7 & $4.51$ & $1.000$ | $1.000$ & $0.992$ | $0.945$ & $0.835$ | $0.377$ & $0.906$ | $0.629$ & $0.696$ | $0.165$ & $0.932$ | $0.787$ & $0.907$ | $0.685$ & $0.885$ | $0.633$ \\
 &  &  &  & 0.25 & 4 & $4.52$ & $1.000$ | $1.000$ & $0.996$ | $0.981$ & $0.892$ | $0.544$ & $0.936$ | $0.746$ & $0.732$ | $0.264$ & $0.926$ | $0.812$ & $0.932$ | $0.732$ & $0.898$ | $0.694$ \\
\cline{4-15}
\noalign{\vskip 0.2ex}
 &  &  & \multirow[t]{2}{*}{5} & 0.50 & 7 & $4.50$ & $1.000$ | $1.000$ & $0.995$ | $0.968$ & $0.882$ | $0.495$ & $0.931$ | $0.705$ & $0.719$ | $0.224$ & $0.940$ | $0.817$ & $0.926$ | $0.708$ & $0.902$ | $0.678$ \\
 &  &  &  & 0.25 & 4 & $4.36$ & $1.000$ | $1.000$ & $0.998$ | $0.985$ & $0.937$ | $0.712$ & $0.964$ | $0.849$ & $0.793$ | $0.380$ & $0.936$ | $0.848$ & $0.961$ | $0.812$ & $0.923$ | $0.766$ \\
\cline{2-15}
\noalign{\vskip 0.2ex}
 & \multirow[t]{20}{*}{+ Clustering} & \multirow[t]{4}{*}{8} & \multirow[t]{2}{*}{2} & 0.50 & 4 & $5.03$ & $0.988$ | $0.969$ & $0.987$ | $0.957$ & $0.959$ | $0.835$ & $0.966$ | $0.863$ & $0.953$ | $0.555$ & $0.892$ | $0.790$ & $0.977$ | $0.905$ & $0.920$ | $0.783$ \\
 &  &  &  & 0.25 & 5 & $5.65$ & $0.995$ | $0.980$ & $0.995$ | $0.973$ & $0.981$ | $0.898$ & $0.972$ | $0.848$ & $0.909$ | $0.394$ & $0.941$ | $0.838$ & $0.989$ | $0.932$ & $0.947$ | $0.796$ \\
\cline{4-15}
\noalign{\vskip 0.2ex}
 &  &  & \multirow[t]{2}{*}{5} & 0.50 & 4 & $6.24$ & $0.999$ | $0.998$ & $0.999$ | $0.997$ & $0.996$ | $0.983$ & $0.992$ | $0.952$ & $0.984$ | $0.765$ & $0.918$ | $0.851$ & $0.997$ | $0.988$ & $0.946$ | $0.873$ \\
 &  &  &  & 0.25 & 5 & $22.08$ & $1.000$ | $1.000$ & $1.000$ | $1.000$ & $0.999$ | $0.995$ & $0.999$ | $0.991$ & $0.977$ | $0.773$ & $0.980$ | $0.935$ & $0.999$ | $0.998$ & $0.984$ | $0.931$ \\
\cline{3-15}
\noalign{\vskip 0.2ex}
 &  & \multirow[t]{4}{*}{16} & \multirow[t]{2}{*}{2} & 0.50 & 2 & $4.58$ & $0.992$ | $0.978$ & $0.988$ | $0.952$ & $0.948$ | $0.780$ & $0.940$ | $0.708$ & $0.988$ | $0.787$ & $0.930$ | $0.830$ & $0.980$ | $0.894$ & $0.945$ | $0.809$ \\
 &  &  &  & 0.25 & 2 & $4.30$ & $0.996$ | $0.990$ & $0.993$ | $0.973$ & $0.951$ | $0.802$ & $0.973$ | $0.890$ & $0.999$ | $0.983$ & $0.948$ | $0.882$ & $0.988$ | $0.951$ & $0.962$ | $0.892$ \\
\cline{4-15}
\noalign{\vskip 0.2ex}
 &  &  & \multirow[t]{2}{*}{5} & 0.50 & 2 & $5.18$ & $1.000$ | $1.000$ & $0.999$ | $0.994$ & $0.987$ | $0.949$ & $0.976$ | $0.874$ & $0.994$ | $0.899$ & $0.953$ | $0.893$ & $0.997$ | $0.983$ & $0.968$ | $0.906$ \\
 &  &  &  & 0.25 & 8 & $9.62$ & $1.000$ | $1.000$ & $1.000$ | $1.000$ & $0.996$ | $0.988$ & $1.000$ | $0.999$ & $0.998$ | $0.981$ & $0.986$ | $0.967$ & $1.000$ | $0.999$ & $0.991$ | $0.977$ \\
\cline{3-15}
\noalign{\vskip 0.2ex}
 &  & \multirow[t]{4}{*}{32} & \multirow[t]{2}{*}{2} & 0.50 & 6 & $4.44$ & $0.991$ | $0.984$ & $0.993$ | $0.979$ & $0.950$ | $0.828$ & $0.996$ | $0.952$ & $0.877$ | $0.477$ & $0.905$ | $0.835$ & $0.982$ | $0.906$ & $0.922$ | $0.805$ \\
 &  &  &  & 0.25 & 6 & $4.55$ & $0.999$ | $0.994$ & $0.997$ | $0.985$ & $0.976$ | $0.898$ & $0.996$ | $0.962$ & $0.949$ | $0.692$ & $0.943$ | $0.870$ & $0.994$ | $0.962$ & $0.959$ | $0.871$ \\
\cline{4-15}
\noalign{\vskip 0.2ex}
 &  &  & \multirow[t]{2}{*}{5} & 0.50 & 6 & $4.54$ & $1.000$ | $1.000$ & $0.999$ | $0.994$ & $0.989$ | $0.953$ & $0.998$ | $0.985$ & $0.920$ | $0.624$ & $0.929$ | $0.880$ & $0.999$ | $0.978$ & $0.946$ | $0.872$ \\
 &  &  &  & 0.25 & 6 & $6.15$ & $1.000$ | $1.000$ & $1.000$ | $0.998$ & $0.998$ | $0.990$ & $0.999$ | $0.993$ & $0.983$ | $0.894$ & $0.974$ | $0.940$ & $1.000$ | $0.998$ & $0.983$ | $0.952$ \\
\cline{3-15}
\noalign{\vskip 0.2ex}
 &  & \multirow[t]{4}{*}{64} & \multirow[t]{2}{*}{2} & 0.50 & 3 & $4.32$ & $0.995$ | $0.988$ & $0.993$ | $0.978$ & $0.935$ | $0.780$ & $0.984$ | $0.890$ & $0.826$ | $0.394$ & $0.904$ | $0.826$ & $0.983$ | $0.896$ & $0.912$ | $0.778$ \\
 &  &  &  & 0.25 & 7 & $4.34$ & $0.999$ | $0.997$ & $0.998$ | $0.990$ & $0.966$ | $0.893$ & $0.987$ | $0.915$ & $0.921$ | $0.563$ & $0.927$ | $0.857$ & $0.993$ | $0.954$ & $0.945$ | $0.841$ \\
\cline{4-15}
\noalign{\vskip 0.2ex}
 &  &  & \multirow[t]{2}{*}{5} & 0.50 & 3 & $4.36$ & $0.999$ | $0.999$ & $0.998$ | $0.996$ & $0.978$ | $0.909$ & $0.994$ | $0.946$ & $0.877$ | $0.520$ & $0.921$ | $0.860$ & $0.997$ | $0.974$ & $0.935$ | $0.841$ \\
 &  &  &  & 0.25 & 7 & $4.85$ & $1.000$ | $1.000$ & $0.999$ | $0.999$ & $0.997$ | $0.989$ & $0.998$ | $0.982$ & $0.962$ | $0.730$ & $0.957$ | $0.911$ & $0.999$ | $0.997$ & $0.972$ | $0.913$ \\
\cline{3-15}
\noalign{\vskip 0.2ex}
 &  & \multirow[t]{4}{*}{128} & \multirow[t]{2}{*}{2} & 0.50 & 4 & $4.34$ & $0.992$ | $0.988$ & $0.991$ | $0.962$ & $0.921$ | $0.712$ & $0.945$ | $0.788$ & $0.804$ | $0.353$ & $0.895$ | $0.807$ & $0.972$ | $0.845$ & $0.902$ | $0.739$ \\
 &  &  &  & 0.25 & 4 & $4.33$ & $1.000$ | $0.997$ & $0.997$ | $0.984$ & $0.956$ | $0.811$ & $0.976$ | $0.866$ & $0.866$ | $0.459$ & $0.928$ | $0.842$ & $0.988$ | $0.901$ & $0.936$ | $0.797$ \\
\cline{4-15}
\noalign{\vskip 0.2ex}
 &  &  & \multirow[t]{2}{*}{5} & 0.50 & 4 & $4.29$ & $1.000$ | $1.000$ & $0.999$ | $0.994$ & $0.972$ | $0.854$ & $0.972$ | $0.866$ & $0.851$ | $0.459$ & $0.916$ | $0.842$ & $0.993$ | $0.928$ & $0.930$ | $0.807$ \\
 &  &  &  & 0.25 & 4 & $4.39$ & $1.000$ | $1.000$ & $1.000$ | $0.996$ & $0.990$ | $0.956$ & $0.994$ | $0.958$ & $0.923$ | $0.683$ & $0.952$ | $0.889$ & $0.999$ | $0.985$ & $0.962$ | $0.888$ \\
\cline{2-15}
\noalign{\vskip 0.2ex}
 & \multirow[t]{4}{*}{+ Cluster Classifier} & \multirow[t]{4}{*}{64} & \multirow[t]{2}{*}{2} & 0.50 & 7 & $4.32$ & $0.994$ | $0.986$ & $0.991$ | $0.927$ & $0.983$ | $0.963$ & $0.977$ | $0.945$ & $0.992$ | $0.983$ & $0.968$ | $0.937$ & $0.971$ | $0.796$ & $0.973$ | $0.920$ \\
 &  &  &  & 0.25 & 7 & $4.34$ & $0.998$ | $0.995$ & $0.994$ | $0.958$ & $0.993$ | $0.977$ & $0.988$ | $0.965$ & $0.998$ | $0.992$ & $0.982$ | $0.961$ & $0.985$ | $0.902$ & $0.985$ | $0.954$ \\
\cline{4-15}
\noalign{\vskip 0.2ex}
 &  &  & \multirow[t]{2}{*}{5} & 0.50 & 3 & $4.36$ & $1.000$ | $1.000$ & $0.996$ | $0.979$ & $0.999$ | $0.996$ & $0.992$ | $0.982$ & $0.999$ | $0.999$ & $0.979$ | $0.964$ & $0.991$ | $0.909$ & $0.985$ | $0.963$ \\
 &  &  &  & 0.25 & 7 & $4.85$ & $1.000$ | $1.000$ & $0.999$ | $0.996$ & $0.999$ | $0.998$ & $0.998$ | $0.997$ & $1.000$ | $1.000$ & $0.993$ | $0.987$ & $0.999$ | $0.989$ & $0.995$ | $0.990$ \\
\cline{1-15}
\noalign{\vskip 0.2ex}
\multirow[t]{32}{*}{RAR-XL} & No WM & - & - & - & - & $3.13$ & $-$ | $-$ & $-$ | $-$ & $-$ | $-$ & $-$ | $-$ & $-$ | $-$ & $-$ | $-$ & $-$ | $-$ & $-$ | $-$ \\
\cline{2-15}
\noalign{\vskip 0.2ex}
 & \multirow[t]{4}{*}{Ours (No Clustering)} & \multirow[t]{4}{*}{-} & \multirow[t]{2}{*}{2} & 0.50 & 3 & $3.06$ & $0.995$ | $0.968$ & $0.963$ | $0.745$ & $0.760$ | $0.153$ & $0.874$ | $0.449$ & $0.697$ | $0.105$ & $0.897$ | $0.618$ & $0.888$ | $0.611$ & $0.855$ | $0.498$ \\
 &  &  &  & 0.25 & 2 & $3.02$ & $0.999$ | $0.989$ & $0.979$ | $0.845$ & $0.788$ | $0.200$ & $0.907$ | $0.551$ & $0.709$ | $0.132$ & $0.914$ | $0.691$ & $0.907$ | $0.661$ & $0.873$ | $0.557$ \\
\cline{4-15}
\noalign{\vskip 0.2ex}
 &  &  & \multirow[t]{2}{*}{5} & 0.50 & 3 & $3.05$ & $1.000$ | $0.995$ & $0.988$ | $0.904$ & $0.809$ | $0.249$ & $0.925$ | $0.634$ & $0.736$ | $0.174$ & $0.922$ | $0.725$ & $0.916$ | $0.693$ & $0.886$ | $0.597$ \\
 &  &  &  & 0.25 & 2 & $3.14$ & $1.000$ | $1.000$ & $0.998$ | $0.974$ & $0.898$ | $0.466$ & $0.969$ | $0.786$ & $0.796$ | $0.299$ & $0.942$ | $0.812$ & $0.949$ | $0.764$ & $0.922$ | $0.700$ \\
\cline{2-15}
\noalign{\vskip 0.2ex}
 & \multirow[t]{20}{*}{+ Clustering} & \multirow[t]{4}{*}{8} & \multirow[t]{2}{*}{2} & 0.50 & 2 & $3.43$ & $0.997$ | $0.958$ & $0.990$ | $0.887$ & $0.931$ | $0.447$ & $0.969$ | $0.721$ & $0.935$ | $0.480$ & $0.955$ | $0.734$ & $0.966$ | $0.716$ & $0.955$ | $0.676$ \\
 &  &  &  & 0.25 & 2 & $3.72$ & $0.997$ | $0.957$ & $0.994$ | $0.917$ & $0.980$ | $0.758$ & $0.967$ | $0.716$ & $0.948$ | $0.626$ & $0.966$ | $0.791$ & $0.983$ | $0.814$ & $0.971$ | $0.771$ \\
\cline{4-15}
\noalign{\vskip 0.2ex}
 &  &  & \multirow[t]{2}{*}{5} & 0.50 & 8 & $4.63$ & $1.000$ | $0.995$ & $0.998$ | $0.975$ & $0.998$ | $0.961$ & $0.977$ | $0.875$ & $0.950$ | $0.705$ & $0.970$ | $0.864$ & $0.996$ | $0.948$ & $0.975$ | $0.868$ \\
 &  &  &  & 0.25 & 2 & $9.72$ & $1.000$ | $1.000$ & $1.000$ | $1.000$ & $1.000$ | $0.994$ & $0.999$ | $0.980$ & $0.999$ | $0.981$ & $0.987$ | $0.955$ & $1.000$ | $0.993$ & $0.997$ | $0.982$ \\
\cline{3-15}
\noalign{\vskip 0.2ex}
 &  & \multirow[t]{4}{*}{16} & \multirow[t]{2}{*}{2} & 0.50 & 1 & $3.05$ & $0.999$ | $0.980$ & $0.990$ | $0.900$ & $0.952$ | $0.585$ & $0.976$ | $0.752$ & $0.916$ | $0.411$ & $0.965$ | $0.789$ & $0.961$ | $0.753$ & $0.961$ | $0.718$ \\
 &  &  &  & 0.25 & 1 & $3.18$ & $0.999$ | $0.986$ & $0.991$ | $0.916$ & $0.913$ | $0.482$ & $0.977$ | $0.789$ & $0.881$ | $0.404$ & $0.960$ | $0.798$ & $0.971$ | $0.802$ & $0.949$ | $0.713$ \\
\cline{4-15}
\noalign{\vskip 0.2ex}
 &  &  & \multirow[t]{2}{*}{5} & 0.50 & 1 & $3.19$ & $1.000$ | $1.000$ & $1.000$ | $0.992$ & $0.991$ | $0.886$ & $0.995$ | $0.944$ & $0.964$ | $0.666$ & $0.981$ | $0.890$ & $0.988$ | $0.890$ & $0.984$ | $0.872$ \\
 &  &  &  & 0.25 & 5 & $5.80$ & $1.000$ | $1.000$ & $1.000$ | $0.997$ & $0.999$ | $0.982$ & $0.996$ | $0.970$ & $0.971$ | $0.753$ & $0.973$ | $0.901$ & $0.997$ | $0.961$ & $0.985$ | $0.912$ \\
\cline{3-15}
\noalign{\vskip 0.2ex}
 &  & \multirow[t]{4}{*}{32} & \multirow[t]{2}{*}{2} & 0.50 & 7 & $3.21$ & $0.998$ | $0.982$ & $0.989$ | $0.894$ & $0.916$ | $0.471$ & $0.938$ | $0.664$ & $0.793$ | $0.243$ & $0.927$ | $0.733$ & $0.951$ | $0.732$ & $0.915$ | $0.643$ \\
 &  &  &  & 0.25 & 2 & $3.08$ & $0.999$ | $0.979$ & $0.993$ | $0.913$ & $0.941$ | $0.478$ & $0.978$ | $0.758$ & $0.897$ | $0.367$ & $0.936$ | $0.730$ & $0.956$ | $0.717$ & $0.943$ | $0.670$ \\
\cline{4-15}
\noalign{\vskip 0.2ex}
 &  &  & \multirow[t]{2}{*}{5} & 0.50 & 7 & $3.31$ & $1.000$ | $1.000$ & $0.999$ | $0.983$ & $0.975$ | $0.769$ & $0.979$ | $0.871$ & $0.872$ | $0.419$ & $0.951$ | $0.838$ & $0.979$ | $0.860$ & $0.951$ | $0.786$ \\
 &  &  &  & 0.25 & 2 & $3.56$ & $1.000$ | $1.000$ & $1.000$ | $0.997$ & $0.994$ | $0.912$ & $0.998$ | $0.981$ & $0.967$ | $0.719$ & $0.967$ | $0.865$ & $0.992$ | $0.903$ & $0.978$ | $0.872$ \\
\cline{3-15}
\noalign{\vskip 0.2ex}
 &  & \multirow[t]{4}{*}{64} & \multirow[t]{2}{*}{2} & 0.50 & 7 & $3.05$ & $0.997$ | $0.975$ & $0.981$ | $0.830$ & $0.815$ | $0.215$ & $0.954$ | $0.652$ & $0.890$ | $0.360$ & $0.910$ | $0.692$ & $0.921$ | $0.669$ & $0.904$ | $0.597$ \\
 &  &  &  & 0.25 & 3 & $3.06$ & $0.998$ | $0.981$ & $0.989$ | $0.911$ & $0.875$ | $0.325$ & $0.953$ | $0.684$ & $0.834$ | $0.276$ & $0.937$ | $0.738$ & $0.946$ | $0.726$ & $0.924$ | $0.636$ \\
\cline{4-15}
\noalign{\vskip 0.2ex}
 &  &  & \multirow[t]{2}{*}{5} & 0.50 & 7 & $3.10$ & $1.000$ | $0.998$ & $0.996$ | $0.966$ & $0.902$ | $0.436$ & $0.984$ | $0.866$ & $0.936$ | $0.514$ & $0.934$ | $0.799$ & $0.962$ | $0.774$ & $0.940$ | $0.728$ \\
 &  &  &  & 0.25 & 3 & $3.96$ & $1.000$ | $1.000$ & $1.000$ | $0.997$ & $0.976$ | $0.766$ & $0.993$ | $0.948$ & $0.929$ | $0.621$ & $0.968$ | $0.868$ & $0.987$ | $0.893$ & $0.972$ | $0.839$ \\
\cline{3-15}
\noalign{\vskip 0.2ex}
 &  & \multirow[t]{4}{*}{128} & \multirow[t]{2}{*}{2} & 0.50 & 2 & $3.06$ & $0.998$ | $0.976$ & $0.979$ | $0.800$ & $0.814$ | $0.191$ & $0.942$ | $0.572$ & $0.793$ | $0.163$ & $0.912$ | $0.665$ & $0.921$ | $0.643$ & $0.892$ | $0.546$ \\
 &  &  &  & 0.25 & 3 & $3.07$ & $0.998$ | $0.988$ & $0.987$ | $0.891$ & $0.865$ | $0.311$ & $0.952$ | $0.659$ & $0.786$ | $0.203$ & $0.919$ | $0.730$ & $0.933$ | $0.699$ & $0.902$ | $0.610$ \\
\cline{4-15}
\noalign{\vskip 0.2ex}
 &  &  & \multirow[t]{2}{*}{5} & 0.50 & 2 & $3.05$ & $0.999$ | $0.996$ & $0.994$ | $0.944$ & $0.893$ | $0.365$ & $0.972$ | $0.775$ & $0.839$ | $0.274$ & $0.934$ | $0.774$ & $0.948$ | $0.729$ & $0.923$ | $0.665$ \\
 &  &  &  & 0.25 & 3 & $3.39$ & $1.000$ | $0.999$ & $0.999$ | $0.992$ & $0.964$ | $0.718$ & $0.987$ | $0.894$ & $0.881$ | $0.466$ & $0.947$ | $0.845$ & $0.973$ | $0.837$ & $0.947$ | $0.788$ \\
\cline{2-15}
\noalign{\vskip 0.2ex}
 & \multirow[t]{4}{*}{+ Cluster Classifier} & \multirow[t]{4}{*}{64} & \multirow[t]{2}{*}{2} & 0.50 & 8 & $3.05$ & $1.000$ | $1.000$ & $0.993$ | $0.954$ & $1.000$ | $0.996$ & $0.997$ | $0.977$ & $1.000$ | $0.997$ & $0.992$ | $0.957$ & $0.937$ | $0.725$ & $0.986$ | $0.931$ \\
 &  &  &  & 0.25 & 1 & $3.06$ & $1.000$ | $0.998$ & $0.996$ | $0.973$ & $0.999$ | $0.994$ & $0.997$ | $0.979$ & $1.000$ | $0.995$ & $0.994$ | $0.967$ & $0.961$ | $0.779$ & $0.990$ | $0.946$ \\
\cline{4-15}
\noalign{\vskip 0.2ex}
 &  &  & \multirow[t]{2}{*}{5} & 0.50 & 6 & $3.10$ & $1.000$ | $1.000$ & $0.998$ | $0.988$ & $1.000$ | $0.999$ & $1.000$ | $0.997$ & $1.000$ | $0.999$ & $0.997$ | $0.982$ & $0.971$ | $0.794$ & $0.994$ | $0.961$ \\
 &  &  &  & 0.25 & 1 & $3.96$ & $1.000$ | $1.000$ & $1.000$ | $0.997$ & $1.000$ | $1.000$ & $1.000$ | $0.999$ & $1.000$ | $1.000$ & $0.998$ | $0.991$ & $0.991$ | $0.923$ & $0.998$ | $0.984$ \\
\bottomrule
\end{tabular}

}
\caption{\textbf{Full experimental results on \textit{Perturbation Set A}} in terms of \textbf{AUC | TPR@FPR=1\% after prefix tuning}.
The reported results use the best performing prefix (from 8 tested prefixes for each setting) in each group where the group is defined by the model, method, number of clusters k, penalty~$\delta$ and green token fraction~$\gamma$. Best prefixes were chosen by taking those with best average TPR@FPR=1\%, averaged across all perturbations (columns).
}
\label{tab:app:full22}
\end{table*}

\begin{table}[]
    \centering
    \scalebox{0.9}{
    \begin{tabular}{lccccccccccc}
\toprule
& +xy=(6,7) & crop=0.11 & crop=0.33 & hflip & prsp=0.3 & rot=33 & rot=7 & scl=0.9 & scl=1.1 & scl=1.5 & tl=0.7 \\
\midrule

&\multicolumn{11}{c}{LlamaGen GPT-B $256\times256$}\\
\midrule
No Clusters        & 0.010 & 0.145 & 0.085 & 0.013 & 0.015 & 0.013 & 0.025 & 0.022 & 0.071 & 0.001 & 0.989 \\
+ Token Classifier & 0.017 & 0.094 & 0.129 & 0.011 & 0.012 & 0.013 & 0.020 & 0.025 & 0.088 & 0.001 & 1.000 \\
\midrule
64 Clusters           & 0.089 & 0.296 & 0.233 & 0.067 & 0.017 & 0.014 & 0.118 & 0.148 & 0.650 & 0.011 & 1.000 \\
+ Token Classifier    & 0.051 & 0.248 & 0.304 & 0.043 & 0.011 & 0.007 & 0.115 & 0.132 & 0.665 & 0.019 & 1.000 \\
+ Cluster Classifier  & 0.064 & 0.332 & 0.276 & 0.059 & 0.009 & 0.003 & 0.143 & 0.141 & 0.593 & 0.025 & 0.999 \\

\midrule
&\multicolumn{11}{c}{LlamaGen GPT-L $384\times384$}\\
\midrule
No Clusters        & 0.013 & 0.135 & 0.151 & 0.020 & 0.017 & 0.011 & 0.042 & 0.049 & 0.224 & 0.004 & 0.999 \\
+ Token Classifier & 0.009 & 0.136 & 0.206 & 0.011 & 0.042 & 0.003 & 0.035 & 0.050 & 0.201 & 0.001 & 0.999 \\
\midrule
64 Clusters           & 0.183 & 0.378 & 0.403 & 0.154 & 0.033 & 0.015 & 0.561 & 0.435 & 0.917 & 0.060 & 0.999 \\
+ Token Classifier    & 0.122 & 0.358 & 0.395 & 0.111 & 0.009 & 0.005 & 0.530 & 0.431 & 0.911 & 0.072 & 1.000 \\
+ Cluster Classifier  & 0.143 & 0.386 & 0.459 & 0.117 & 0.012 & 0.004 & 0.529 & 0.422 & 0.844 & 0.099 & 1.000 \\

\midrule
&\multicolumn{11}{c}{RAR-XL $256\times256$}\\
\midrule

No Clusters        & 0.035 & 0.221 & 0.216 & 0.024 & 0.025 & 0.013 & 0.069 & 0.111 & 0.576 & 0.003 & 0.987 \\
+ Token Classifier & 0.025 & 0.406 & 0.498 & 0.021 & 0.064 & 0.011 & 0.205 & 0.336 & 0.972 & 0.003 & 1.000 \\
\midrule

64 Clusters          & 0.062 & 0.405 & 0.404 & 0.033 & 0.026 & 0.012 & 0.177 & 0.312 & 0.838 & 0.009 & 0.996 \\
+ Token Classifier   & 0.059 & 0.533 & 0.615 & 0.043 & 0.077 & 0.008 & 0.495 & 0.616 & 0.995 & 0.013 & 1.000 \\
+ Cluster Classifier & 0.038 & 0.465 & 0.526 & 0.040 & 0.051 & 0.005 & 0.365 & 0.547 & 0.989 & 0.011 & 1.000 \\

\bottomrule

\end{tabular}
}
    \caption{With geometric transformations: Detection accuracy (TPR@FPR=1\%) with our proposed methods (fixed $\gamma=0.25$ and $\delta=5$), deployed with LlamaGen (GPT-B and GPT-L) and RAR-XL.
    The geometric transformations are: (1) +xy: translation, (2) crop by a factor, effectively lowering resolution, (3) horizontal flip, (4) prsp: perspective transform, (5) rot: rotation by a degree, (6) scl: scaling by a factor while cropping if the image is larger than original, (7) crop\_tl: cropping the top-left part of the image by a factor.
    }
    \label{tab:geom}
\end{table}

\begin{table}[]
    \centering
    \scalebox{0.9}{
    \begin{tabular}{lccccccccc}
\toprule
 & rot=33 & rot=7 & scl=0.5 & scl=0.7 & scl=0.9 & scl=1.1 & scl=1.5 \\

\midrule
&\multicolumn{7}{c}{LlamaGen GPT-B $256\times256$}\\
\midrule
No Clusters        & 0.988 & 0.989 & 0.884 & 0.988 & 0.994 & 0.995 & 0.995 \\
+ Token Classifier & 1.000 & 1.000 & 0.997 & 0.999 & 1.000 & 1.000 & 1.000 \\
\midrule
64 Clusters           & 1.000 & 1.000 & 0.997 & 1.000 & 1.000 & 1.000 & 1.000 \\
+ Token Classifier    & 1.000 & 1.000 & 1.000 & 1.000 & 1.000 & 1.000 & 1.000 \\
+ Cluster Classifier  & 0.999 & 1.000 & 1.000 & 1.000 & 1.000 & 1.000 & 1.000 \\

\midrule
&\multicolumn{7}{c}{LlamaGen GPT-L $384\times384$}\\
\midrule
No Clusters        & 1.000 & 1.000 & 0.982 & 0.998 & 0.999 & 1.000 & 1.000 \\
+ Token Classifier & 1.000 & 1.000 & 0.999 & 1.000 & 1.000 & 1.000 & 1.000 \\
\midrule
64 Clusters           & 1.000 & 1.000 & 1.000 & 1.000 & 1.000 & 1.000 & 1.000 \\
+ Token Classifier    & 1.000 & 1.000 & 1.000 & 1.000 & 1.000 & 1.000 & 1.000 \\
+ Cluster Classifier  & 1.000 & 1.000 & 1.000 & 1.000 & 1.000 & 1.000 & 1.000 \\

\midrule
&\multicolumn{7}{c}{RAR-XL $256\times256$}\\
\midrule

No Clusters        & 0.994 & 0.990 & 0.956 & 0.998 & 0.999 & 1.000 & 1.000 \\
+ Token Classifier & 1.000 & 1.000 & 1.000 & 1.000 & 1.000 & 1.000 & 1.000 \\
\midrule

64 Clusters           & 0.998 & 0.998 & 0.993 & 1.000 & 1.000 & 1.000 & 1.000 \\
+ Token Classifier    & 1.000 & 1.000 & 1.000 & 1.000 & 1.000 & 1.000 & 1.000 \\
+ Cluster Classifier  & 1.000 & 1.000 & 1.000 & 1.000 & 1.000 & 1.000 & 1.000 \\

\bottomrule
\end{tabular}
}
    \caption{With reverted geometric transformations: Detection accuracy (TPR@FPR=1\%) with our proposed methods (fixed $\gamma=0.25$ and $\delta=5$), deployed with LlamaGen (GPT-B and GPT-L) and RAR-XL.}
    \label{tab:geom:undo}
\end{table}

\begin{table}[]
    \centering
    \scalebox{0.9}{
    \begin{tabular}{lcccccccc}
\toprule
 & hflip & prsp=0.3 & rot=33 & rot=7 & scl=0.7 & scl=1.5 & scl=1.5 + gaussian blur R=3 & SyncSeal \\

\midrule
&\multicolumn{8}{c}{LlamaGen GPT-B $256\times256$}\\
\midrule
No Clusters        & 0.893 & 0.927 & 0.917 & 0.943 & 0.703 & 0.966 & 0.358 & 0.989 \\
+ Token Classifier & 0.947 & 0.986 & 0.977 & 0.985 & 0.919 & 0.992 & 0.858 & 0.996 \\
\midrule
64 Clusters           & 0.991 & 0.997 & 0.990 & 0.997 & 0.985 & 0.997 & 0.936 & 0.999 \\
+ Token Classifier    & 0.997 & 1.000 & 0.998 & 0.999 & 0.998 & 0.999 & 0.997 & 1.000 \\
+ Cluster Classifier  & 0.991 & 0.997 & 0.984 & 0.997 & 0.996 & 0.997 & 0.982 & 1.000 \\

\midrule
&\multicolumn{8}{c}{LlamaGen GPT-L $384\times384$}\\
\midrule
No Clusters        & 0.849 & 0.916 & 0.970 & 0.988 & 0.553 & 0.988 & 0.630 & 0.999 \\
+ Token Classifier & 0.888 & 0.932 & 0.986 & 0.994 & 0.725 & 0.996 & 0.953 & 1.000 \\
\midrule
64 Clusters           & 0.989 & 0.994 & 0.999 & 1.000 & 0.933 & 1.000 & 0.990 & 1.000 \\
+ Token Classifier    & 0.992 & 0.996 & 1.000 & 1.000 & 0.951 & 1.000 & 1.000 & 1.000 \\
+ Cluster Classifier  & 0.987 & 0.992 & 0.995 & 1.000 & 0.932 & 1.000 & 0.996 & 1.000 \\

\midrule
&\multicolumn{8}{c}{RAR-XL $256\times256$}\\
\midrule

No Clusters & 0.912 & 0.959 & 0.841 & 0.921 & 0.858 & 0.979 & 0.376 & 0.994 \\
+ Token Classifier & 0.995 & 0.998 & 0.993 & 0.997 & 0.992 & 0.997 & 0.998 & 0.999 \\
\midrule

64 Clusters & 0.971 & 0.987 & 0.937 & 0.983 & 0.945 & 0.993 & 0.661 & 1.000 \\
+ Token Classifier & 1.000 & 1.000 & 0.996 & 0.999 & 0.997 & 1.000 & 1.000 & 1.000 \\
+ Cluster Classifier & 0.999 & 1.000 & 0.991 & 0.999 & 0.995 & 1.000 & 0.999 & 1.000 \\

\bottomrule
\end{tabular}
}
    \caption{With SyncSeal-reverted geometric transformations: Detection accuracy (TPR@FPR=1\%) with our proposed methods (fixed $\gamma=0.25$ and $\delta=5$), deployed with LlamaGen (GPT-B and GPT-L) and RAR-XL.}
    \label{tab:geom:syncseal}
\end{table}


\end{document}